\documentclass[a4paper,fleqn]{cas-dc}

\usepackage[numbers]{natbib}

\usepackage{hyperref}
\usepackage{csquotes}
\usepackage[most]{tcolorbox}
\usepackage{array}
\newcolumntype{L}[1]{>{\raggedright\arraybackslash}p{#1}}
\usepackage{booktabs}
\usepackage{amsmath}

\usepackage{listings}
\def\tsc#1{\csdef{#1}{\textsc{\lowercase{#1}}\xspace}}
\tsc{WGM}
\tsc{QE}
\tsc{EP}
\tsc{PMS}
\tsc{BEC}
\tsc{DE}

\begin{document}
\let\WriteBookmarks\relax
\def\floatpagepagefraction{1}
\def\textpagefraction{.001}
\shorttitle{Detecting Soft Skills in ML Engineering Roles CVs}
\shortauthors{A. Azamnouri et~al.}

\title [mode = title]{Detecting Soft Skills in ML Engineering Roles CVs}

\author[1]{Aidin Azamnouri}[orcid=0009-0004-8409-2316]
\cormark[1]
\ead{aidin.azamnouri@tum.de}

\credit{Conceptualization of this study, Methodology, Software}

\affiliation[1]{organization={Chair of Software Engineering, TUM School of Computation, Information and Technology, Technical University of Munich},
                city={Heilbronn},
                country={Germany}}

\author[1]{Nouran Ayad}

\author[2]{Justus Bogner}[orcid=0000-0001-5788-0991]

\affiliation[2]{organization={Software and Sustainability (S2) Group, Vrije Universiteit Amsterdam},
                city={Amsterdam},
                country={The Netherlands}} 

\author[1]{Stefan Wagner}[orcid=0000-0002-5256-8429]

\cortext[cor1]{Corresponding author}

\begin{abstract}
Soft skills shape collaboration among ML engineers, data scientists, and software engineers building ML-enabled systems, yet what we know about them comes almost entirely from the demand side. Job advertisements, surveys, and hiring manager interviews capture what employers ask for. How candidates themselves articulate these competencies has not been studied, and existing CV-mining work is both keyword-based, so it cannot see skills conveyed through narrative, and descriptive, reporting frequency rankings without testing whether group differences exceed sampling variation. We close both gaps. Using a balanced corpus of 300 curated CVs spanning the three roles, we extract explicitly listed and implicitly narrated soft skills with an LLM-based pipeline validated against a human-annotated ground truth, a distinction that existing extractors were not designed to make. We then convert the demand-side literature's claims into 13 falsifiable hypotheses about role signatures, seniority progression, and disclosure style, and test them with effect sizes under family-wise error control, so that candidate-side data can corroborate or contradict the demand-side account rather than merely illustrate it. Eleven hypotheses are supported, one partially, and one refuted. Candidates disclose soft skills through narrative rather than keyword lists by roughly three to one, and most so for the competencies employers value most: leadership, coordination, and mentoring (88–96\% narrative). Seniority nearly triples the odds of articulating leadership. That competency, assumed universal in prior work, is articulated by software engineers at half the rate of their peers. Technical candidates do articulate soft skills, in a form keyword-based screening systematically misses.
\end{abstract}



\maketitle

\section{Introduction}

The development of machine learning (ML)-enabled systems increasingly demands collaboration and communication across multiple engineering roles, including software engineers, ML engineers, and data scientists~\cite{Nahar_2022, Azamnouri_2025, Haug_2025}. While these roles differ in their technical responsibilities, they share a common dependency. Successful delivery requires effective collaboration and communication among team members, not just the correct implementation of the requirements and needs~\cite{Nahar_2022, Mailach_2023, Busquim_2024}. As ML systems become more deeply embedded in business-critical products, organizations face challenges that are not purely technical. Hence, these challenges make soft skills an increasingly important complement to technical expertise in modern artificial intelligence (AI) and software engineering work~\cite{Mohammed_2024, De_Morais_Le_a_2025}.

Soft skills have been repeatedly recognized as key predictors of career success and project effectiveness~\cite{Patacsil_2017, Deming_2017, Mohammed_2024, Romanenko_2024}. Prior work suggests that while technical expertise provides the baseline for employability, long-term professional success is heavily shaped by soft skills such as leadership, collaboration, and communication~\cite{Patacsil_2017, Piorkowski_2021, Vu_2026}. In highly interdisciplinary domains like AI, these competencies become even more critical because the work is inherently distributed~\cite{Nahar_2022, Busquim_2024_2, Azamnouri_2025}. ML practitioners must translate model outcomes into actionable insights, justify trade-offs to product teams, and collaborate with software and Dev\-Ops engineers to ensure maintainability and robustness of deployed pipelines~\cite{Haug_2025}. As a result, job postings for ML-related roles increasingly emphasize soft skills alongside technical qualifications~\cite{Vu_2026}. Despite this demand, the industry continues to report a persistent soft-skills gap, particularly among early-career applicants, raising concerns about how technical professionals are trained, evaluated, and hired~\cite{Nahar_2022, Busquim_2024, Romanenko_2024}. At the same time, understanding soft skill expectations and workforce readiness remains challenging. Most empirical studies that examine soft skills in engineering labor markets rely on job postings or survey instruments, which capture employer intent rather than candidate self-presentation~\cite{Calanca_2019, Succi_2019, Lyu_2021, Kovacs_2022}. Another major body of work focuses on resume mining and automated skill extraction, yet these approaches often depend on dictionary-based keyword matching~\cite{Gugnani_2020, Wosiak_2021, Wang_2021, Sinha_2021}. Such methods are brittle and fail to capture the nuanced and varied ways candidates describe soft skills in real-world CVs (Curriculum Vitae).

Despite the centrality of soft skills to collaboration in ML-enabled system development, most empirical evidence to date has been drawn from job postings or surveys, which reflect employer expectations rather than how candidates themselves articulate these competencies. As a result, relatively little is known about how ML engineers, data scientists, and software engineers (three important roles for building ML-enabled systems) actually present soft skills in their CVs, how this presentation varies across roles, and how it evolves with seniority. Understanding the candidate side of this picture matters for several reasons. For hiring managers and recruiters, it reveals whether the soft skills demanded in job postings are actually surfaced in applications, and where screening pipelines built on keyword matching may systematically miss competencies that candidates convey through narrative rather than explicit labels. For practitioners, particularly early-career ML and MLOps engineers facing the well-documented soft skills gap, it provides concrete evidence of how experienced professionals in the same role communicate collaboration, leadership, and mentoring, offering a reference point for their own self-presentation. For educators and curriculum designers, role-specific and seniority-specific patterns indicate which competencies are most visibly rewarded in practice and where training programs may need to place stronger emphasis. And for researchers studying the socio-technical dimensions of ML engineering, CV-side evidence complements the existing job-posting literature by showing whether the collaboration, communication, and coordination needs identified in prior work on ML-enabled system development are reflected in how practitioners describe their own contributions.

Crucially, the demand-side literature does not merely describe employer preferences; it advances explanatory claims about \textit{why} particular competencies matter for particular roles and career stages. Data scientists are described as occupying a translational position between analytical results and business stakeholders~\cite{Nahar_2022, Busquim_2024_2}; ML engineers as bridging software engineering and data science teams~\cite{Nahar_2022, Haug_2025}; senior roles as demanding strategic competencies that junior roles do not~\cite{Peretz_2025}; and leadership as a universal rather than role-specific expectation in software work~\cite{Galster_2022, Malinen_2025}. Each of these claims implies a testable prediction about what candidate-authored documents should look like. Treating those predictions as hypotheses rather than only as interpretive framing is what allows a candidate-side study to corroborate or contradict the demand-side account, rather than merely illustrate it. We therefore state the predictions explicitly, test them inferentially, and report which survive.

This paper addresses these gaps through an empirical study of 300 professionals' CVs spanning the three roles. To enable analysis at scale, we develop a supporting generative information-extraction pipeline based on a large language model (LLM). Compared with traditional dictionary-based or rule-based extraction methods, generative models are better suited to interpreting the contextual and often implicit ways in which soft skills are expressed in CVs~\cite{Mohamed_2025, Harnad_2025}. We treat this pipeline as a methodological enabler rather than the central contribution of the paper. It makes comparative analysis tractable across 300 heterogeneous documents, but the empirical findings on role and seniority differences are ultimately what this work is about.
In summary, this paper makes the following contributions:
\begin{enumerate}
\item \textbf{A confirmatory empirical analysis of soft skill articulation} across ML engineers, data scientists, and software engineers. We derive 13 falsifiable hypotheses from the demand-side literature, covering role signatures, seniority progression, and disclosure style, and test them with inferential statistics under family-wise error control, reporting odds ratios, risk differences, and standardized effect sizes with 95\% confidence intervals. We complement this with false-discovery-controlled exploratory contrasts across the full taxonomy, and with sensitivity analyses that quantify how much extraction error would be required to overturn each conclusion.
\item \textbf{A reproducible curation methodology} including role-specific filtering rules, a tri-party annotation protocol for resolving role ambiguity, duplicate-detection procedures, and the search queries used to source contemporary CVs from public web indexes. We apply this methodology to construct a balanced corpus of 300 CVs by combining a previously established multi-labeled CV corpus with a supplementary set of contemporary CVs sourced from publicly accessible sources.
\item \textbf{A supporting LLM-based extraction pipeline} that elicits both explicit and implicit soft skill mentions from unstructured CV narratives, validated iteratively against a human-annotated ground truth. We include this pipeline to make the analysis tractable and reproducible at scale, not as a contribution to the skill-extraction literature.
\end{enumerate}

Our findings offer actionable guidance for practitioners preparing CVs for ML-related roles, educators designing professional development curricula, and researchers studying workforce readiness in AI-intensive fields.

The remainder of this paper is structured as follows. Section~\ref{sec:related-work} reviews related work on soft skills in software engineering and automated CV analysis. Section~\ref{sec:methodology} describes our research questions, hypotheses, dataset, LLM-based extraction methodology, and statistical analysis plan. Section~\ref{sec:validation-pipeline} presents the evaluation design of the extraction approach. Section~\ref{sec:results} reports results across role, seniority, and rhetorical patterns, including the outcomes of all hypothesis tests and the accompanying robustness analyses. Section~\ref{sec:discussion} discusses the results and implications, while Section~\ref{sec:threats-to-validity} describes threats to validity. Section~\ref{sec:conclusion} concludes and outlines future research directions.

\section{Related Work}
\label{sec:related-work}

A substantial body of research demonstrates that soft skills are not peripheral but central to professional success in software and AI-intensive environments~\cite{Niva_2020, Piorkowski_2021, Nahar_2022, Mailach_2023, Azamnouri_2025}. Empirical analyses of job advertisements consistently reveal that employers explicitly demand interpersonal and behavioral competencies alongside technical expertise. For instance, a large-scale study of software engineering job advertisements in New Zealand found that more than four-fifths of postings explicitly required soft skills, with communication emerging as the most frequently cited requirement across company sizes and position types~\cite{Galster_2022}. Beyond job advertisements, qualitative investigations provide complementary evidence. Through in-depth interviews with practicing software professionals, Malinen et al.~\cite{Malinen_2025} report strong consensus regarding the importance of communication, teamwork, and leadership, while also highlighting less visible yet critical competencies such as resilience and self-awareness. Together, these findings underscore that soft skills are not merely symbolic additions in job descriptions but reflect genuine workplace expectations.

Skill requirements also evolve substantially across career stages. Research examining AI professionals across junior and senior levels shows a clear shift in emphasis. Early-career roles prioritize foundational programming and analytical capabilities, whereas senior roles increasingly require strategic competencies, including leadership, stakeholder communication, and business-oriented thinking~\cite{Peretz_2025}. This transition suggests that soft skills become progressively central as professionals assume greater responsibility, influence, and cross-functional coordination. Importantly, hiring manager perspectives further illuminate a discrepancy between candidate preparation and employer expectations. An exploratory case study conducted within software engineering recruitment contexts~\cite{Loufek_2025} found that hiring managers often prioritize non-technical competencies over specific technical tools when evaluating entry-level applicants. Collectively, these studies motivate the need for systematic, data-driven analyses of how soft skills are actually articulated in CVs.

A separate line of work has developed computational methods for identifying skills in unstructured recruitment text. Early resume parsing frameworks relied on rule-based pipelines and Named Entity Recognition~\cite{Sajid_2022}, which handled clearly delineated fields such as education and listed skills but struggled with implicitly expressed competencies. Ontology-based systems~\cite{Phan_2021} and syntax-aware extractors~\cite{Smith_2019} extended this work by modeling semantic relationships and leveraging dependency structure, respectively, while the SkillSpan dataset~\cite{Zhang_2022} established Transformer-based benchmarks for span-level hard and soft skill annotation in job postings. More recent systems, such as SkillGPT~\cite{Li_2023}, combine LLM-based summarization with embedding similarity to map skills onto structured taxonomies. We build on this lineage rather than adopt any of these systems directly, because our study imposes a requirement that they were not designed to meet, which is how a candidate disclosed soft skills, explicitly or implicitly. Existing extractors target the former and are not constructed to separate listed competencies from narrated ones. We therefore implement a generative extraction pipeline that elicits both forms and ties each extracted skill to its supporting phrase, enabling the explicit-versus-implicit distinction our analysis depends on. The pipeline is a means to that analysis rather than a contribution to skill extraction in its own right.

Taken together, prior work establishes that soft skills matter, that employer demand for them is well-documented, and that increasingly capable tools exist to extract them from text. What remains underexplored is the candidate side of the picture. Three gaps are particularly salient. First, existing empirical analyses of soft skills in technical labor markets~\cite{Calanca_2019, Galster_2022, Malinen_2025, Peretz_2025, Loufek_2025} draw almost exclusively on job advertisements, survey instruments, or practitioner interviews; none systematically examine how candidates themselves articulate soft skills in their CVs. Second, the comparative dimension is missing. While differences between junior and senior AI professionals have been studied at the level of job-posting requirements~\cite{Peretz_2025}, there is no comparable evidence on how soft skill articulation varies across ML engineers, data scientists, and software engineers, or how it evolves with seniority, when observed directly from candidate-authored documents. Third, and methodologically, the CV-mining literature is overwhelmingly descriptive. Studies report frequency rankings of extracted skills but rarely test whether the differences between groups exceed what sampling variation would produce, and rarely report effect sizes that would allow the practical importance of a difference to be judged. This paper addresses these gaps through a hypothesis-driven analysis of 300 CVs spanning the three roles, examining role-specific emphasis, seniority-based progression, and the rhetorical patterns candidates use to present soft skills.

\section{Methodology}
\label{sec:methodology}

\begin{figure*}
\centerline{\includegraphics[width=0.8\textwidth]{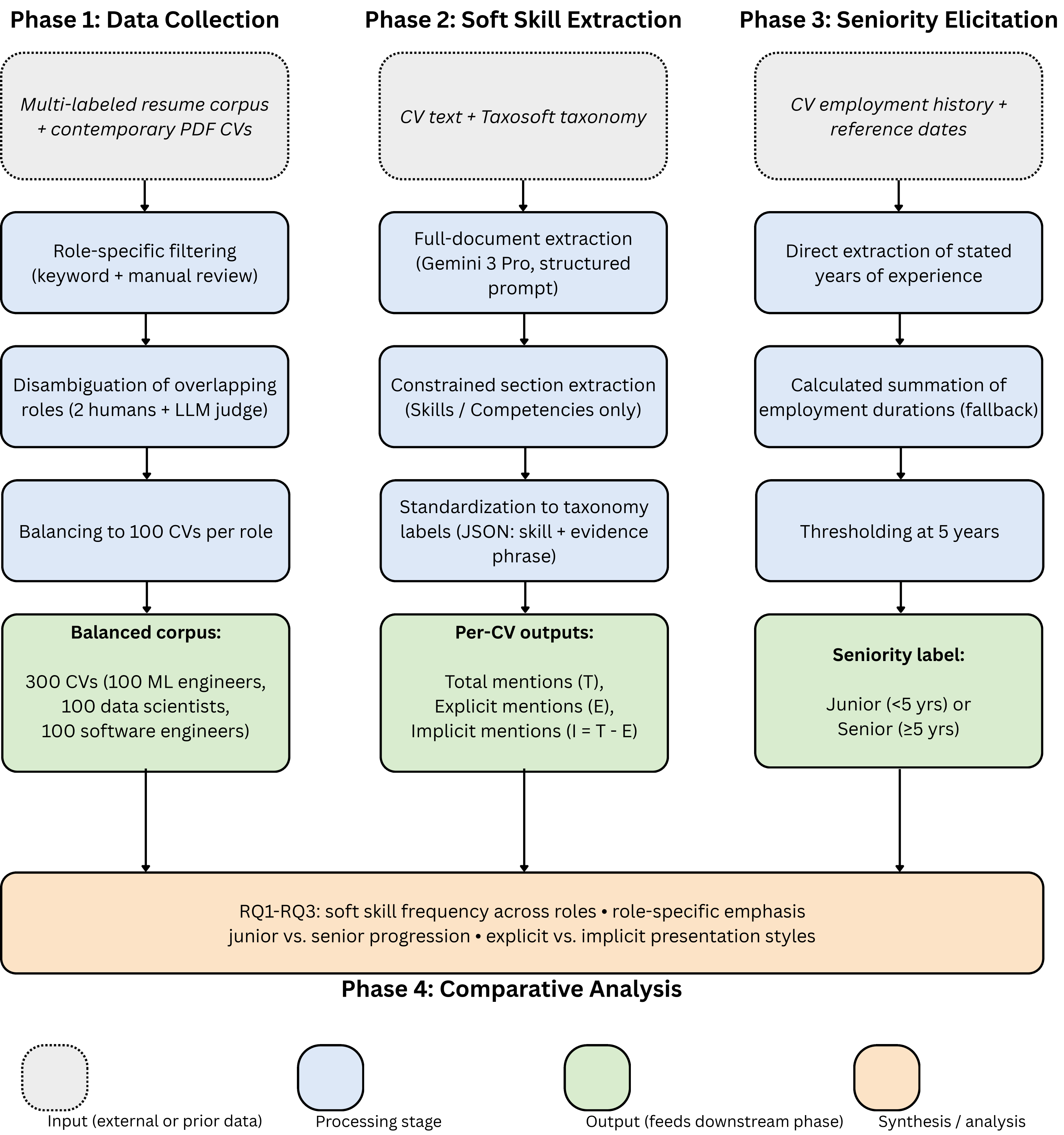}}
\caption{An overview of the study process}
\label{fig1}
\end{figure*}

This section presents the research design and technical framework for examining how soft skills are mentioned in professional CVs. To overcome the limitations of traditional keyword-based parsing approaches, this study adopts a generative information extraction strategy that leverages LLMs' semantic reasoning capabilities. In particular, models from the Gemini family are employed to interpret unstructured CV narratives and identify behavioral competencies embedded in descriptions of professional experience and achievements. The methodological approach is supported by a hybrid data collection strategy that combines a previously established multi-labeled CV corpus with a supplementary dataset of contemporary PDF CVs. This combination ensures both methodological robustness and temporal relevance, enabling the analysis to capture both established patterns and emerging trends in how professionals present soft skills in technical CVs. The steps of the study process are depicted in Fig.~\ref{fig1}.

To address the gap in the literature, this paper aims to answer the following research questions:

\textbf{RQ1}: What soft skills are most commonly mentioned by ML engineers, software engineers, and data scientists, and how does the emphasis on these skills differ across the three roles?

This question characterizes the overall landscape of soft skill articulation in the dataset and identifies role-specific patterns. Answering it reveals which behavioral competencies are most visible to recruiters across ML-related CVs and highlights where each role places distinctive emphasis.

\textbf{RQ2}: How do soft skills mentions vary by seniority level?

This question examines whether and how the articulation of soft skills evolves as professionals advance in their careers. Comparing junior and senior candidates reveals whether CVs shift from emphasizing collaborative participation to leadership, coordination, and strategic competencies as seniority increases.

\textbf{RQ3}: Do candidates present soft skills explicitly as keywords in dedicated skills sections, or implicitly through narrative descriptions of their professional experience?

This question investigates the rhetorical patterns candidates use to present soft skills, distinguishing between explicit mentions, in which skills are listed as keywords in dedicated sections, and implicit mentions, in which behavioral competencies are conveyed through narrative descriptions of professional experience. The answer indicates whether candidates rely primarily on keyword-based self-labeling or storytelling-based disclosure, and how this balance varies across roles and seniority levels.

\subsection{Hypotheses}
\label{sec:hypotheses}

The three research questions above are stated as open questions because no prior study has examined soft skill articulation from the candidate side. Prior work on the demand side, however, does license specific expectations about what candidate-side data should look like if the mechanisms proposed in that literature operate as described. We therefore translate each research question into a set of falsifiable hypotheses, so that our analysis can confirm or disconfirm those expectations rather than only describe frequencies. Table~\ref{tab:hypotheses} summarizes the hypotheses, their grounding in prior work, and the statistical test used for each; their outcomes are reported in Table~\ref{tab:hypresults}.

We distinguish two classes of hypotheses. \textit{Confirmatory} hypotheses are derived from published findings that predate our analysis; for these, we fix the comparison and the direction in advance and control the family-wise error rate. \textit{Exploratory} contrasts are the remaining comparisons, which we report with false-discovery-rate control and interpret as hypothesis-generating. We note explicitly that the hypotheses were derived from the literature but were not preregistered, and that all tests are performed on a single sample; the confirmatory label therefore refers to derivation from prior theory, not to preregistration. The distinction is nonetheless consequential for interpretation. The confirmatory set fixes 13 comparisons in advance out of the several hundred that the taxonomy would permit, so that the analysis can fail, whereas the exploratory set is explicitly framed as generating claims for future studies to test.

\textbf{Role signatures (RQ1).} Studies of ML-enabled system development describe data scientists as occupying a translational position between analytical results and business stakeholders~\cite{Nahar_2022, Busquim_2024_2}, and communication is the single most frequently demanded competency in software job advertisements~\cite{Galster_2022}. If this translational burden is reflected in self-presentation, it should be visible as a role difference in communication.

\begin{itemize}
\item \textbf{H\textsubscript{1a}:} Data scientists mention communication more frequently than ML engineers and software engineers.
\item \textbf{H\textsubscript{1b}:} ML engineers mention mentoring more frequently than data scientists and software engineers.
\item \textbf{H\textsubscript{1c}:} The proportion of CVs with no detectable soft skills differs across roles.
\item \textbf{H\textsubscript{1d}:} Data scientists frame problem-solving as a behavioral competency more frequently than ML engineers.
\item \textbf{H\textsubscript{1e}:} Leadership prevalence does not differ across the three roles.
\end{itemize}

Because H\textsubscript{1e} predicts the absence of an effect, a non-significant omnibus test would be uninformative on its own. Failure to reject the null is not evidence for it. We therefore accompany the omnibus test with two one-sided equivalence tests (TOST) on each pairwise risk difference, using an equivalence margin of $\pm$20 percentage points. The margin is set to the smallest role difference we consider practically meaningful for the applications this paper addresses. A gap below 20 percentage points would not change how a curriculum or a screening pipeline is designed for one role versus another, whereas a gap above it would. H\textsubscript{1e} is treated as supported only if the omnibus test is non-significant and all pairwise contrasts are statistically equivalent within that margin.

\textbf{Seniority progression (RQ2).} Prior analysis of junior and senior AI job advertisements~\cite{Peretz_2025} reports that senior roles demand strategic competencies (leadership, stakeholder communication, business orientation) that junior advertisements do not, while junior advertisements prioritize foundational technical skills. An exploratory study of software engineering recruitment~\cite{Loufek_2025} reports that hiring managers treat non-technical competencies as decisive at the entry level even though candidates rarely surface them. If these demand-side patterns have candidate-side counterparts, the following should hold.

\begin{itemize}
\item \textbf{H\textsubscript{2a}:} Senior professionals mention leadership more frequently than junior professionals.
\item \textbf{H\textsubscript{2b}:} Senior professionals mention coordination more frequently than junior professionals.
\item \textbf{H\textsubscript{2c}:} Senior professionals mention strategic thinking more frequently than junior professionals.
\item \textbf{H\textsubscript{2d}:} The proportion of CVs with no detectable soft skills is lower among senior than among junior professionals.
\item \textbf{H\textsubscript{2e}:} Two competing accounts make opposite predictions for collaboration. A \textit{substitution} account predicts that collaboration is mentioned less often by senior professionals as leadership displaces it. An \textit{accumulation} account predicts that senior professionals mention collaboration at least as often as they do.
\item \textbf{H\textsubscript{2f}:} The seniority effect on leadership is not an artifact of role composition.
\end{itemize}

\textbf{Disclosure style (RQ3).} A substantial share of CV screening infrastructure matches keywords against predefined skill vocabularies~\cite{Gugnani_2020, Wosiak_2021}, which presupposes that candidates label competencies explicitly. Qualitative work reports instead that soft skills are vaguely described and inconsistently framed in candidate documents~\cite{Malinen_2025, Loufek_2025}.

\begin{itemize}
\item \textbf{H\textsubscript{3a}:} Within the same CV, soft skills are disclosed through narrative more often than through explicit labels in a dedicated skills section.
\item \textbf{H\textsubscript{3b}:} Disclosure style depends on role, with ML engineers relying on narrative disclosure most heavily.
\end{itemize}

\begin{table*}[t]
\centering
\caption{Confirmatory hypotheses, their grounding in prior work, and the statistical test applied to each.}
\label{tab:hypotheses}
\small
\begin{tabular}{|l|p{5.6cm}|l|p{4.6cm}|}
\hline
\textbf{ID} & \textbf{Prediction} & \textbf{Grounding} & \textbf{Test} \\
\hline
H\textsubscript{1a} & Communication: data scientists $>$ ML + software engineers & \cite{Nahar_2022, Busquim_2024_2, Galster_2022} & Fisher's exact, $2\times2$ \\
\hline
H\textsubscript{1b} & Mentoring: ML engineers $>$ data scientists + software engineers & \cite{Nahar_2022, Haug_2025} & Fisher's exact, $2\times2$ \\
\hline
H\textsubscript{1c} & No-skill rate differs across roles & \cite{Nahar_2022} & Pearson $\chi^2$, $3\times2$; planned contrast \\
\hline
H\textsubscript{1d} & Problem solving: data scientists $>$ ML engineers & \cite{Nahar_2022} & Fisher's exact, $2\times2$ \\
\hline
H\textsubscript{1e} & Leadership: no difference across the three roles & \cite{Galster_2022, Malinen_2025} & Pearson $\chi^2$, $3\times2$ + pairwise TOST ($\pm$20 pp) \\
\hline
H\textsubscript{2a} & Leadership: senior $>$ junior (primary hypothesis) & \cite{Peretz_2025} & Fisher's exact, $2\times2$ \\
\hline
H\textsubscript{2b} & Coordination: senior $>$ junior & \cite{Peretz_2025} & Fisher's exact, $2\times2$ \\
\hline
H\textsubscript{2c} & Strategic thinking: senior $>$ junior & \cite{Peretz_2025} & Fisher's exact, $2\times2$ \\
\hline
H\textsubscript{2d} & No-skill rate: senior $<$ junior & \cite{Loufek_2025, Romanenko_2024} & Fisher's exact, $2\times2$ \\
\hline
H\textsubscript{2e} & Collaboration: substitution vs.\ accumulation (non-directional) & \cite{Peretz_2025} & Fisher's exact, $2\times2$ \\
\hline
H\textsubscript{2f} & Seniority effect on leadership survives adjustment for role and is homogeneous across roles & \cite{Peretz_2025} & Logistic regression (LR test), CMH, Breslow--Day \\
\hline
H\textsubscript{3a} & Within a CV, narrative disclosure $>$ keyword disclosure & \cite{Gugnani_2020, Wosiak_2021, Malinen_2025} & Exact McNemar; Wilcoxon signed-rank \\
\hline
H\textsubscript{3b} & Disclosure style depends on role; ML engineers most narrative-reliant & \cite{Loufek_2025, Malinen_2025} & Pearson $\chi^2$, $3\times4$; planned contrast \\
\hline
\end{tabular}
\end{table*}

\subsection{Data Collection}

The foundational dataset for this study is the multi-labeled CV corpus developed by Jiechieu and Tsopze~\cite{Jiechieu_2020}, which is publicly available on GitHub and has been used for benchmarking NLP tasks on professional documents. The corpus provides three properties that are particularly valuable for the present work: (1) full-text CVs in plain \textit{.txt} format, which allow direct computational processing without the need for optical character recognition; (2) expert-labeled occupation tags, which serve as a ground-truth reference for role-based analysis; and (3) a normalization file that maps raw job titles to a smaller set of standardized professional classes.

Although the corpus contains 19,465 CVs spanning a broad range of occupations, its built-in normalization file was found to be unsuitable for this study. In that file, the expert curators collapsed both \enquote{ML Engineer} and \enquote{Data Scientist} into a single broad \enquote{Software Developer} category, which would make it impossible to distinguish the three target roles of this work. The normalized file was therefore discarded in favor of the raw archive, in which each CV is stored as a three-field record separated by the \enquote{:::} delimiter: a reference identifier, a semicolon-separated list of raw occupation titles as written by domain experts, and the full text of the CV. Working from the raw occupation list rather than the normalized classes allowed us to design a more granular filtering procedure tailored to the three roles of interest.
All filtering steps described below were implemented as case-insensitive queries, applied either to the raw occupation list or to the full CV text, depending on the step.

\textbf{ML engineers.} A first pass retrieved all CVs whose text matched the pattern \enquote{ML} or \enquote{machine learning}, yielding 204 candidates. Because the token \enquote{ML} is highly polysemous in technical CVs, i.e., it frequently appears as a listed tool or library rather than as a job title, a second pass required the stricter pattern \enquote{ML } (with a trailing whitespace) so that matches would fire only when \enquote{ML} prefixed another word, as in \enquote{ML Engineer}. This reduced the candidate pool to 115 CVs. Each of these was then subjected to a manual heuristic review designed to confirm that ML development was the candidate's primary occupation rather than a secondary activity. The review consisted of a qualitative assessment of three CV components:
\begin{enumerate}
    \item \textit{Professional summary or abstract.} Candidate profiles were evaluated based on self-identification in their introductory narrative. A CV was retained only if the summary explicitly described the individual as an \enquote{ML Engineer} or used a synonymous technical title, e.g., \enquote{Machine Learning Engineer}, \enquote{AI/ML Engineer}, \enquote{MLOps Engineer}, \enquote{Deep Learning Engineer}, or \enquote{Applied Machine Learning Engineer}.
    \item \textit{Work experience.} The professional history of each candidate was examined to verify that their primary or most recent roles involved the core development, deployment, or maintenance of ML models. Candidates whose experience suggested only incidental use of ML tools within non-engineering roles, e.g., data analysts applying pre-built models for reporting, were excluded.
    \item \textit{Technical skills.} A distinction was drawn between ML as a tool and ML as a core competency. This involved verifying whether listed technical skills, such as Scikit-learn or TensorFlow, were supported by evidence of participation in the full ML lifecycle, including feature engineering, model evaluation, and MLOps practices, rather than appearing as a bare list of library names.
\end{enumerate}

\textbf{Data scientists.} A single-pass keyword filter for the job title \enquote{Data Scientist} identified 105 CVs. Each CV was then manually reviewed using the same three-component protocol described above, with the retention criterion adapted to the data science role. This review yielded 94 usable profiles.

\textbf{Software engineers.} A broader initial filter for \enquote{Software Developer} and \enquote{Software Engineer} returned 5,322 CVs, reflecting the size and heterogeneity of this role in the corpus. To balance dataset size across roles and ensure higher relevance, the search was narrowed to candidates who listed \enquote{Software Engineer} as their primary (first) occupation, reducing the set to 363 CVs. Finally, to prevent overlap with the ML and data-science subsets, any CV whose text also matched the ML or data-science patterns above was removed, leaving 357 CVs for the software engineering subset.

Because the Jiechieu and Tsopze corpus was compiled in 2020~\cite{Jiechieu_2020}, it does not reflect the most recent shifts in how technical professionals describe themselves in CVs. To bridge this five-year gap and bring the dataset in line with current labor-market practices, a supplementary dataset of contemporary CVs was manually curated.

For that, publicly hosted CVs were collected through targeted Google queries using the \textit{filetype:pdf} operator to restrict results to PDF documents. Then, they were manually checked to remove the sample ones. The specific queries used were:
\begin{itemize}
    \item ML CV filetype:pdf and Machine Learning CV filetype:pdf
    \item data scientist CV filetype:pdf
    \item software engineer CV filetype:pdf
\end{itemize}

\begin{figure*}
\centerline{\includegraphics[width=0.90\textwidth]{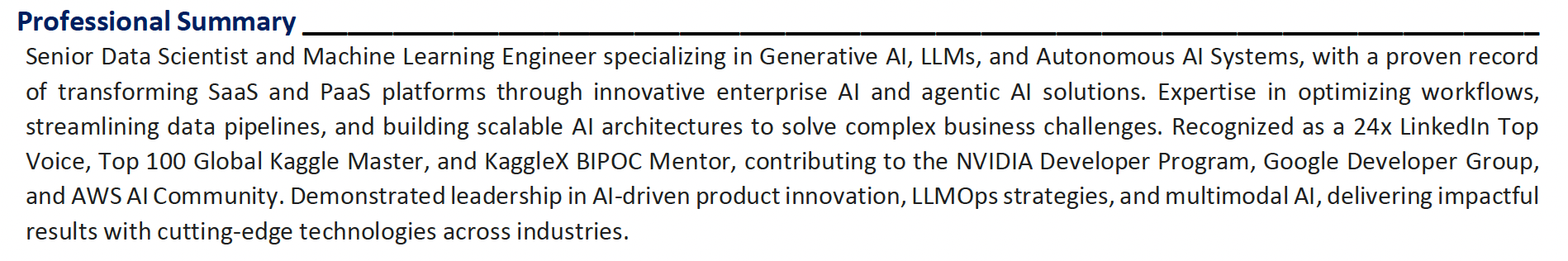}}
\caption{An example of ML engineer/data scientist role ambiguity in a CV}
\label{fig2}
\end{figure*}

In contrast to the GitHub corpus, where role assignment relied on expert-labeled occupation tags, supplementary CVs were selected based on the role explicitly stated in the candidate's document title or header. Documents were retained only when (1) the file represented a genuine individual CV rather than a template, sample, or job posting, and (2) the role declared in the title or header matched one of the three target roles. Some profiles remained ambiguous after this initial pass, particularly at the boundary between data science and ML engineering, and were resolved through the classification procedure described below.

Thirteen CVs were identified in which candidates explicitly self-identified as belonging to both the data science and ML engineering roles. This ambiguity is illustrated in Fig.~\ref{fig2}, which shows a professional summary in which a candidate simultaneously describes themselves as a \enquote{Senior Data Scientist and Machine Learning Engineer}. To assign these CVs to a mutually exclusive role, a voting mechanism was employed involving two human researchers and an LLM (Gemini 2.5 Pro) acting as a third annotator. Each annotator independently classified the 13 disputed profiles, and the final assignment was determined by majority vote.

The decision to include an LLM as a supplementary annotator is motivated by a growing body of empirical evidence~\cite{Croxford_2025}. Recent work in software engineering research~\cite{Ahmed_2025, Wang_2025, He_2025} has shown that mixed human--LLM evaluation designs can achieve inter-rater agreements comparable to human-only settings, particularly when high-performance models are used, which tend to reflect a majority opinion distilled from their training corpora~\cite{Ahmed_2025}. The use of a 2-to-1 majority voting rule for difficult cases similarly aligns with established practice for subjective and nuanced annotation tasks, where even professional raters frequently disagree. Current guidance in the literature is that LLMs should not yet fully replace human annotators in software engineering studies, but can serve as effective supplementary annotators that reduce human effort while maintaining data reliability~\cite{Ahmed_2025}. The tri-party design adopted here follows the recommendation that the LLM never act as the sole arbiter and that no case be decided by the LLM alone, as each outcome requires agreement from at least one human annotator.

To validate the classification procedure, the degree of consensus among the three annotators was analyzed for the 13 overlapping cases. In 9 of the 13 cases (69.2\%), all three annotators independently converged on the same role, indicating professional narratives that, despite surface-level ambiguity, were substantively aligned with a single role. In the remaining 4 cases (30.8\%), a 2-to-1 majority was required; these CVs typically described hybrid profiles with significant technical overlap between the two roles. The LLM's assignment aligned with the final majority decision in 11 of 13 cases (84.6\%), indicating strong agreement with the aggregated human consensus without dominating it. The two cases in which the LLM diverged from the final classification were both resolved by the two human annotators agreeing against the LLM, confirming that human judgment remained the decisive factor.

To enable a fair comparative analysis across the three roles, the dataset was balanced to include exactly 100 CVs per role. The choice of 100 CVs per role reflects an upper bound on what was practically feasible to curate at the level of quality required for this study, while remaining sufficient to support the comparative analyses; Section~\ref{sec:stats} reports the resulting sensitivity of the design in terms of the smallest differences it can reliably detect. Three practical constraints shaped this decision. First, every CV in the dataset was subject to manual review by the same two human researchers who carried out the role disambiguation procedure described above. As described in the role-specific filtering and disambiguation steps above, this review involved reading the professional summary, the most recent work experience entries, and the listed competencies for each candidate, both to confirm the primary occupation and to verify narrative completeness. Scaling this human-in-the-loop process beyond a few hundred CVs would have compromised the consistency of role assignment, which is a prerequisite for the comparative claims this paper makes. Second, the supply of suitable contemporary CVs was itself very limited. After filtering out templates, sample CVs published by career-services websites, job postings, duplicates, non-English documents, and CVs whose declared role did not unambiguously match one of the three target roles, the pool of remaining publicly accessible CVs was effectively saturated. We also made concerted efforts to solicit CVs from the MLOps community by clearly communicating our research objectives; however, no submissions were received from community members. Additional queries with broader keywords returned fewer new, high-quality candidates and a higher proportion of already collected or out-of-scope documents, indicating that the relevant public web corpus had been substantially covered. Third, 100 CVs per group is consistent with sample sizes used in prior empirical studies of soft-skill articulation in technical labor-market documents~\cite{Khaouja_2019}, thereby supporting comparability with the existing literature.

For the ML engineering subset, the 47 validated profiles from the GitHub corpus were combined with 53 manually collected contemporary CVs, reaching the target of 100. For the data science and software engineering subsets, where the corpus pool was substantially larger than required, a purposeful manual selection was applied instead of random sampling: 62 CVs were selected for data science and 53 for software engineering, based on text completeness. Text completeness refers to CVs that contain the full set of expected sections (summary, experience, skills, education). These selections were then combined with the contemporary supplementary set, which added 38 data science CVs and 47 software engineering CVs, bringing the total to 100 CVs per role. The final balanced corpus comprises 300 CVs and serves as the foundation for the comparative analyses presented in the remainder of this paper.

\begin{table*}[t]
\centering
\caption{Final balanced dataset composition}
\label{tab1}
\begin{tabular}{|c|c|c|c|}
\hline
\textbf{Profession} & \textbf{GitHub Corpus (2020)} & \textbf{Manual (2025)} & 
\textbf{Total Analyzed}\\
\hline

ML Engineer & 47 & 53 & 100 \\
\hline
Data Scientist & 62 & 38 & 100 \\
\hline
Software Engineer & 53 & 47 & 100 \\
\hline
\textbf{Total} & \textbf{162} & \textbf{138} & \textbf{300} \\
\hline
\end{tabular}
\end{table*}

To prevent accidental duplication between the corpus-derived and contemporary subsets, each contemporary CV collected via the queries was checked against the GitHub corpus prior to inclusion. Specifically, for each candidate's CV, we computed normalized text overlap against all corpus CVs of the same target role using a token-level Jaccard similarity and fuzzy matching over the professional summary and the most recent two work experience entries; any candidate exceeding a similarity threshold of 0.5 was manually inspected, and exact or near-exact matches were excluded. No duplicates were identified through this procedure. The final dataset composition is shown in Table~\ref{tab1}.

All CVs analyzed in this study originate from publicly accessible sources, and the analysis uses them solely for aggregate, role-level statistical purposes; no individual candidate is named or identified in any artifact released alongside this paper. The CVs themselves are not redistributed, both because a subset of the contemporary documents include explicit author statements against republication and because anonymizing publicly indexed CVs would offer no meaningful privacy protection while leaving the originals traceable through web search.

\subsection{Soft Skill Extraction and Standardization}

After constructing the balanced dataset, a structured extraction pipeline was implemented to convert raw CV text into a standardized set of soft skill categories. The pipeline relies on a generative information extraction approach in which an LLM processes full CV narratives and identifies behavioral competencies expressed within them. In contrast to traditional NLP pipelines, no extensive preprocessing techniques such as lemmatization or stop-word removal were applied. This design choice was motivated by the observation that modern generative models perform more effectively when provided with complete contextual sentences rather than fragmented tokens~\cite{Xu_2024}. For CVs originally stored in PDF format, the PyMuPDF (fitz)\footnote{\url{https://pymupdf.readthedocs.io/en/latest/}} library was used to extract raw textual content. This popular library was selected because it preserves the reading order of textual blocks, allowing the model to correctly interpret relationships between sections such as job titles, role descriptions, and listed competencies.

Soft skill identification was performed using models from the Gemini family. Each CV was provided to the model as a complete text document accompanied by a structured prompt\footnote{can be found in \url{https://doi.org/10.5281/zenodo.21873447}} instructing the model to identify and extract soft skills present in the candidate's professional narrative. The model's task was to detect both explicit skill mentions and implicit behavioral competencies and map them to a predefined taxonomy of soft skills, TaxoSoft~\cite{Khaouja_2019}. This taxonomy was developed to address the terminology gap commonly observed in professional documents, where equivalent competencies are described using varied language, e.g., \enquote{Collaboration} versus \enquote{Teamwork}. TaxoSoft constructs a structured representation of soft skills by combining knowledge graph resources such as DBpedia\footnote{\url{https://www.dbpedia.org/}} with distributional semantics derived from Word2Vec\footnote{\url{https://code.google.com/archive/p/word2vec/}}. The resulting taxonomy defines relationships between skills using hierarchical structures derived from social network analysis, where central skills act as parent nodes and related competencies are represented as subordinate nodes, e.g., \enquote{Communication} is a parent to \enquote{Active Listening} and \enquote{Verbal Communication}. This extraction strategy follows the generative information extraction paradigm, which represents a shift away from traditional sequence labeling methods~\cite{Xu_2024}. In this paradigm, LLMs directly generate structured information from unstructured input, enabling more flexible and context-aware extraction in real-world textual environments where rigid patterns are insufficient.

Several factors motivated the use of LLM-based extraction for this task. First, generative models support zero-shot and few-shot generalization, allowing them to perform complex information extraction tasks using only natural language instructions and minimal examples. This capability reduces the dependence on large task-specific labeled datasets, which are often unavailable for CV analysis. Second, LLMs can simultaneously address multiple extraction subtasks, such as identifying entities and interpreting contextual relationships, within a single unified framework~\cite{Decorte_2023}. This capability is particularly valuable in CV parsing, which is an extreme multi-label problem in which numerous overlapping competencies may appear within a single document. Furthermore, unlike rule-based systems that rely on rigid pattern matching, LLMs leverage semantic knowledge acquired during pretraining, enabling them to distinguish between a skill listed as a keyword and a competency demonstrated through professional experience. Finally, LLM-based extraction offers significant scalability advantages compared with manual annotation processes, which are time-consuming and costly when applied to large text corpora.

To facilitate downstream analysis, the LLM was instructed to return extraction results in a structured JSON format. The schema required the model to produce a list of skill objects containing two fields: a standardized skill label derived from the taxonomy and the exact phrase from the CV that signaled the presence of the corresponding competency (see listing~\ref{lst1}). Including the original textual phrase alongside the normalized skill label improves interpretability and enables verification of the model's reasoning. This design allows us to confirm whether an identified skill is genuinely supported by textual evidence rather than resulting from model hallucination. Moreover, comparing extracted phrases with their assigned categories provided valuable feedback during iterative prompt refinement, thereby improving extraction accuracy. To maintain consistency across the dataset, an additional verification step by the first and second author ensured that all extracted competencies corresponded to entries within the selected taxonomy. When the model produced alternative linguistic variants, these were automatically mapped to their canonical taxonomy labels, thereby standardizing the representation of soft skills across all CVs in the dataset.

\begin{lstlisting}[caption={An example of structured json output format for skill extraction},label={lst1}]
[
 {
   "skill": "Collaboration",
   "phrase": "Collaborated with cross-functional teams"
 },
 {
   "skill": "Communication",
   "phrase": "Maintained positive communication with product owners"
 }
]
\end{lstlisting}

\subsection{Seniority Level Elicitation}

To analyze how soft skill articulation evolves across career stages, each CV in the dataset was assigned a seniority label. Industry hiring standards for advanced technical roles, particularly in data science and ML, often require approximately 5 years of relevant professional experience before candidates are considered for senior-level responsibilities~\cite{indeed_senior_ds_berlin, glassdoor_senior_ds_berlin, levels_fyi_levels, robert_half_2026}. These positions typically involve technical execution, project coordination, mentoring, stakeholder interaction, and competencies that are rarely developed within less than a five-year timeframe. Consequently, to better reflect industry practice and improve statistical balance, this study adopts a five-year threshold to determine seniority. Under this framework, professionals with fewer than five years of relevant experience are categorized as junior, whereas those with five or more years are classified as senior. Although professional job titles vary across organizations, e.g., associate, lead, or principal, this binary classification provides a consistent analytical baseline for examining changes in soft skill expression across career progression. Because dichotomizing a continuous variable discards information and can distort effect estimates, we additionally retain the underlying estimate of total years of experience and use it in a robustness analysis (Section~\ref{sec:robustness}).

To accomplish this, an automated extraction procedure was implemented using the Gemini 3 Pro Preview model. The model was instructed to determine a candidate's experience level through two complementary strategies. First, it attempted direct extraction by identifying explicit statements indicating total years of professional experience within sections such as the professional summary or profile. When such information was unavailable, the model applied a calculated summation approach, identifying employment start and end dates for each listed position and computing the cumulative duration of the candidate's professional experience. This approach grounds the seniority classification in the chronological timeline of the CV rather than relying on potentially ambiguous titles.

To maintain temporal accuracy when calculating experience durations, the extraction pipeline incorporated fixed reference dates corresponding to the time at which each dataset was collected. This step was particularly necessary for roles marked as \enquote{Present} where the end date of employment is unspecified. For CVs originating from the GitHub corpus, which was collected on June 20, 2020, ongoing positions were calculated relative to this date. For the supplementary CVs gathered through targeted web searches, a reference date of December 1, 2025 was used as an approximation of the data collection period. By combining experience-based inference with explicit temporal anchors, the methodology ensures a consistent and reproducible classification of junior and senior professionals across the dataset.

\subsection{Soft Skill Presentation Styles}

In addition to identifying which soft skills appear in CVs, this study also examines how candidates present these skills. Specifically, the methodology distinguishes between skills explicitly listed as keywords and those implicitly conveyed through narrative descriptions of professional experience. To enable this analysis, a two-stage extraction strategy was implemented to separate structurally declared competencies from those inferred through contextual storytelling.

First, a constrained section extraction procedure was applied to isolate skills that candidates explicitly highlight. Using the Gemini 3 Pro Preview model, the extraction prompt was restricted to specific CV segments typically dedicated to competency summaries. The model was instructed to identify soft skills only within sections labeled Skills, Technical Competencies, or Core Strengths. At the same time, the model was explicitly prevented from extracting information from other narrative sections such as Work Experience, Professional History, Projects, or Education. This restriction ensures that the extracted skills represent deliberate self-labeling by the candidate, typically expressed as keywords or bullet-point lists.

The second stage focuses on implicit skill identification. During the initial extraction phase of the pipeline, the model processed the full CV text and extracted all identifiable soft skills, including those embedded within descriptions of tasks, achievements, and responsibilities. These comprehensive results constitute the total set of detected skills. To distinguish narrative-based skills from explicitly declared ones, a deductive comparison was applied between the two extraction outputs. Skills obtained from the constrained skills-section extraction were categorized as explicit mentions ($E$), while the complete set extracted from the entire document was defined as total mentions ($T$). The set of implicit mentions ($I$) was then derived by subtracting explicit mentions from the total set ($I = T - E$). This analytical procedure enables the quantification of different soft skill disclosure styles within professional CVs. For example, if a competency, such as leadership, appears in descriptions of project coordination or team management but is absent from the candidate's skills section, it is classified as an implicit mention. Such cases indicate that the candidate demonstrates the skill through professional narratives rather than directly labeling it. By separating these two forms of expression, the methodology allows for a systematic analysis of the balance between keyword-based self-presentation and storytelling-based skill communication across roles and seniority levels. Because $E$ and $I$ are disjoint by construction and jointly exhaust $T$, the two disclosure channels can be compared within the same CV as a paired observation, which is what H\textsubscript{3a} requires.

\subsection{Statistical Analysis}
\label{sec:stats}

This subsection specifies the analysis plan applied to the extracted data. All analyses are conducted at the level of the individual CV. For each CV and each taxonomy label, we record a binary indicator of whether that competency was detected, so that a candidate who describes collaboration in four separate bullet points contributes one positive observation rather than four. This de-duplication makes the unit of analysis the candidate rather than the sentence, which is the appropriate level for claims about how professionals present themselves. It yields 1,354 unique skill--CV pairs from the 1,367 raw mentions extracted by the pipeline; of these pairs, 347 are disclosed as keywords in a dedicated skills section and 1,007 through narrative. Frequencies reported as percentages throughout Section~\ref{sec:results} are therefore the proportion of CVs in a group in which a competency was detected at least once, not counts of mentions.

\textbf{Tests.} For $2\times2$ comparisons of a binary outcome between two independent groups, we use Fisher's exact test, which remains valid for the sparse cells that arise for lower-frequency competencies, for example, strategic thinking appears in only 6 junior CVs. For omnibus comparisons across the three roles, and for the $3\times4$ comparison of disclosure style by role, we use Pearson's $\chi^2$ test without continuity correction; expected cell counts were verified to be adequate ($\ge 3.3$ in the sparsest cell of the disclosure-style table, and $\ge 8$ elsewhere). H\textsubscript{3a} compares two measurements made on the same CV and is therefore tested with the exact McNemar test on the discordant pairs, complemented by a Wilcoxon signed-rank test on the per-CV counts of implicitly versus explicitly disclosed competencies. H\textsubscript{2f} is tested with logistic regression. We compare nested models by likelihood-ratio test to assess the seniority effect adjusted for role, and the seniority~$\times$~role interaction to assess homogeneity. As a non-parametric check on the same question, we report the Cochran--Mantel--Haenszel (CMH) common odds ratio stratified by role, together with the Breslow--Day test of homogeneity of odds ratios across strata. Differences in the number of competencies per CV are tested with Kruskal--Wallis (role) and Mann--Whitney $U$ (seniority), since these counts are right-skewed and zero-inflated. All tests are two-sided, including those attached to directional hypotheses, so that a result in the direction opposite to the prediction can be detected rather than absorbed into a non-significant tail.

\textbf{Effect sizes.} Reporting significance without magnitude would leave the practical question unanswered, so every test is accompanied by an effect size with a 95\% confidence interval. For $2\times2$ tables, we report the conditional maximum-likelihood odds ratio with its exact confidence interval, the risk difference in percentage points with a Newcombe hybrid-score interval, and Cohen's $h$. For omnibus $\chi^2$ tests, we report Cramér's $V$. For the rank-based tests, we report the rank-biserial correlation and $\epsilon^2$. Proportions reported on their own carry Wilson score intervals.

\textbf{Multiplicity.} The 13 confirmatory hypotheses in Table~\ref{tab:hypotheses} form a single family, and we control the family-wise error rate across them at $\alpha=0.05$ using the Holm--Bonferroni procedure. Where a hypothesis includes a planned follow-up contrast (H\textsubscript{1c}, H\textsubscript{1e}, H\textsubscript{3b}), the contrast is corrected separately within its own small family. The exploratory analyses test every competency detected in at least 15 CVs (24 labels) against role and against seniority; these form two further families, each controlled at a 5\% false discovery rate using the Benjamini--Hochberg procedure. We report both raw and adjusted $p$-values throughout, and we do not describe an exploratory result as a finding unless it survives adjustment.

\textbf{Sensitivity of the design.} The sample size was fixed by curation constraints rather than by a power calculation, so we report the resulting sensitivity rather than a post-hoc power estimate. At $\alpha=0.05$ and 80\% power, against a baseline prevalence of 30\%, the design can detect a difference of 19.2 percentage points between two roles (100 vs.\ 100), 16.6 points between one role and the other two combined (100 vs.\ 200), and 16.1 points between senior and junior professionals (186 vs.\ 114). Differences smaller than these are not reliably resolvable in this corpus, which is why we report confidence intervals for every contrast and why H\textsubscript{1e} is evaluated by equivalence testing rather than by a non-significant $p$-value.

\textbf{Robustness.} Three sensitivity analyses accompany the confirmatory results and are reported in Section~\ref{sec:robustness}. First, because the seniority variable is a dichotomization, we re-estimate the primary hypothesis using the underlying continuous estimate of years of experience. Second, because longer CVs offer more opportunities for a competency to be detected, we re-estimate the primary hypothesis, adjusting for document length. Third, and most importantly, because the outcome variable is produced by an imperfect extractor, we conduct a quantitative bias analysis that propagates the pipeline's measured sensitivity and specificity through the estimates, and we compute how large a differential extraction error between groups would have to be before each conclusion reverses.

\section{Validation of the Extraction Pipeline}
\label{sec:validation-pipeline}

This section describes the evaluation framework used to assess the effectiveness of the proposed soft skill extraction approach. Because the empirical claims in Section~\ref{sec:results} are only as reliable as the extraction that produces them, this section validates the pipeline against a human-annotated ground truth before applying it to the full corpus. The measured precision and recall reported here are not only a quality statement about the pipeline; they are inputs to the quantitative bias analysis in Section~\ref{sec:robustness}, which propagates them through every confirmatory estimate. The evaluation relies on standard information extraction metrics, precision, recall, and F1-score, to measure the accuracy and completeness of the extracted skills~\cite{Goutte_2005}. These metrics are widely adopted in information extraction and skill mining research and provide a balanced assessment of system performance~\cite{Gugnani_2018, Khaouja_2019, Kolluru_2020, Akkasi_2024}. In this study, the F1-score serves as the primary evaluation metric because it simultaneously penalizes false positives (hallucinated skills not present in the text) and false negatives (skills present in the CV but not detected by the model). Consequently, it provides a balanced measure of both precision, the proportion of correctly extracted skills, and recall, the proportion of relevant skills successfully identified. This evaluation framework enables direct comparison with prior research in skill extraction and information extraction systems.

To ensure rigorous assessment, the development of the extraction pipeline followed an iterative refinement process rather than a single implementation step. The system was progressively improved through 22 experimental trials organized into four developmental phases, tracking the transition from an initial baseline configuration to a highly optimized prompt-based extraction system. This iterative design was necessary due to the non-deterministic behavior of generative AI models and the linguistic complexity of CVs, which frequently contain ambiguous phrasing, technical noise, and varied terminology for similar competencies. Each iteration focused on refining the prompt structure, improving taxonomy alignment, and introducing additional constraints to reduce hallucinations while preserving high recall.

In addition, a human-validated ground truth dataset was constructed to provide a reliable benchmark for evaluating each experimental iteration. This dataset consists of a randomly selected sample of 100 CVs drawn from the original corpus. The size of this evaluation subset aligns with prior research on skill extraction systems~\cite{Khaouja_2019}, ensuring methodological comparability. During the annotation process, each CV was manually reviewed by human annotators who identified soft skills based exclusively on the labels and alternative labels defined in the adopted TaxoSoft taxonomy~\cite{Khaouja_2019}. For every identified skill, the annotators also recorded the exact phrase in the CV that signaled the presence of that competency. These phrase--skill pairs served as the gold standard for evaluating the extraction results generated by the LLM. During the annotation process, each of the 100 CVs was independently reviewed by the first and second authors, who carried out the role disambiguation procedure. Both annotators worked from the same instructions and identified soft skills based exclusively on the labels and alternative labels defined in the adopted TaxoSoft taxonomy~\cite{Khaouja_2019}.

Given the exploratory and interpretive nature of the annotation task, we did not compute a formal inter-coder reliability statistic. Instead, reliability was ensured through repeated joint calibration and consensus-based resolution of disagreements, which are widely used strategies in qualitative research to ensure analytical rigor~\cite{Guest_2012, Saldana_2021}. Conflicts between the two annotators fell into three categories, each handled by an explicit rule: (1) cases in which one annotator identified a skill that the other had not noted were resolved through joint re-examination of the CV, with the skill retained only if both annotators agreed that the supporting phrase met the taxonomy's criteria; (2) cases in which both annotators identified the same skill but cited different supporting phrases were resolved by retaining the phrase judged to most directly express the competency, with ties broken by selecting the earlier occurrence in the CV; and (3) cases in which the two annotators assigned different taxonomy labels to the same phrase were resolved through discussion, referring back to the TaxoSoft alternative-label definitions to determine the canonical category. The reconciled phrase--skill pairs served as the gold standard for evaluating the extraction results generated by the LLM.

Following each experimental trial, the system's JSON output was automatically compared with the ground truth dataset. Precision, recall, and F1-score were calculated by comparing the extracted skill--phrase pairs with the annotated reference set. Any discrepancies were carefully analyzed to identify the underlying causes of false positives and false negatives. These analyses informed subsequent prompt adjustments, such as introducing stronger extraction constraints, refining taxonomy variants, and incorporating negative examples to prevent misclassifying technical tasks as behavioral competencies. This feedback loop ensured that each prompt revision resulted in measurable improvements in extraction accuracy.

\begin{table*}[htbp]
\caption{Evolution of extraction performance across all refinement phases. F1 is the harmonic mean of the precision and recall values computed before rounding.}
\centering
\begin{tabular}{|l|r|r|r|l|}
\hline
\textbf{Phase / Milestone} & \textbf{Precision} & \textbf{Recall} & \textbf{F1-score} & \textbf{Key Refinements / Models} \\
\hline
Baseline (Trial 1) & 0.32 & 0.60 & 0.41 & Flash-Lite (v2.5); initial taxonomy sample \\
\hline
Scaling (Trial 6) & 0.31 & 0.84 & 0.45 & Gemini 2.5 Flash; architecture upgrade \\
\hline
Constrained (Trial 8) & 0.35 & 0.82 & 0.49 & Gemini 2.5 Flash; negative examples; variant pruning \\
\hline
Optimized (Trial 12) & 0.63 & 0.80 & 0.70 & Gemini 2.5 Pro; explicit grounding rules; final refinement \\
\hline
\textbf{Peak Performance (Trial 13)} & \textbf{0.69} & \textbf{0.74} & \textbf{0.72} & \textbf{Gemini 3 Pro Preview} \\
\hline
\end{tabular}
\label{tab2}
\end{table*}

\begin{table*}[htbp]
\caption{Comparative Benchmarking of Soft Skill Extraction Performance}
\centering
\begin{tabular}{|l|l|l|r|}
\hline
\textbf{Study/Methodology} & \textbf{Data Source} & \textbf{Approach} & \textbf{F1-score}\\
\hline
This Study (Optimized Gemini) & CVs & Generative Information Extraction & 0.72 \\
\hline
Akkasi~\cite{Akkasi_2024} & Job Ads & Transformer Ensemble & 0.67 \\
\hline
TaxoSoft~\cite{Khaouja_2019} & Job Ads & Hybrid Dictionary & 0.84 \\
\hline
Knowledge Base (Baseline)~\cite{Malherbe_2016} & Job Ads & Wikipedia/DBpedia & 0.54 \\
\hline
ESCO Taxonomy (Baseline)~\cite{De_2015} & Job Ads & Manual/Keyword & 0.17 \\
\hline
\end{tabular}
\label{tab3}
\end{table*}

The refinement process progressed through four major phases. The progression of performance metrics across these phases is summarized in Table~\ref{tab2}. The first phase established the baseline configuration and integrated the TaxoSoft taxonomy into the prompt structure. Early experiments relied on simple natural language instructions and partial taxonomy samples, which resulted in relatively low performance. Subsequent trials introduced few-shot examples and expanded the taxonomy coverage, improving the model's ability to recognize skill--phrase relationships. The second phase focused on constraint engineering, introducing explicit exclusion rules and negative examples to reduce hallucinated extractions from technical descriptions. The third phase involved model scaling and prompt optimization, transitioning to larger Gemini architectures and applying automated prompt optimization techniques to improve extraction consistency. The final phase conducted cross-model benchmarking, evaluating a variety of large-scale open-source and proprietary language models, e.g., openai-gpt-oss-120b, deepseek-r1-0528, and llama-3.1-sauerkrautlm-70b-instruct, using the optimized prompt configuration. Despite testing several high-capacity architectures, the tuned Gemini 3 Pro Preview configuration achieved the highest performance, reaching a precision of 0.69, a recall of 0.74, and an F1-score of 0.72.

The selection of comparison models was guided by three considerations. First, all benchmarked models were available through the institutional API infrastructure used in this study, which provided consistent access conditions and reproducible inference settings across configurations. Second, the comparison set was deliberately weighted toward open-weight models, which together cover the largest publicly inspectable architectures available at the time of the study and align with the recommendation of Baltes et al.~\cite{Baltes_2025} to include open LLM baselines in empirical software engineering studies. Third, the comparison was designed to test whether the optimized prompt generalized across model families rather than to benchmark every available frontier proprietary model. Models such as OpenAI's GPT-5 and Anthropic's Claude Opus 4 were not included in this comparison because they were not part of the institutional API access available during the study period; we view their evaluation as a natural direction for future work, particularly as their pricing and access conditions evolve.

Finally, to confirm that extraction quality is adequate for the comparative analysis that follows, we situate the pipeline's performance relative to prior soft skill extraction approaches (Table~\ref{tab3}). Notably, the proposed generative information extraction approach achieves higher performance than several traditional baselines while effectively addressing the terminology gap inherent in unstructured CVs.
While the TaxoSoft framework~\cite{Khaouja_2019} reported a higher F1-score, several methodological differences explain this gap. First, the TaxoSoft system employed a hybrid methodology combining DBpedia and Word2Vec to build a comprehensive dictionary of exact-match terms and morphological variants for each skill category. Second, the TaxoSoft evaluation was conducted on job advertisements, where recruiters typically use standardized professional keywords to improve searchability and visibility. In contrast, CVs contain far more diverse and personalized language, making skill identification considerably more challenging. As highlighted by Xu et al.~\cite{Xu_2024}, generative information extraction from highly unstructured documents introduces additional complexity due to varied formatting and narrative expression. Also, the current methodology emphasized rigorous evaluation by relying on explicit textual grounding, deliberately limiting skill extraction to those directly supported by phrases to reduce the hallucination issues commonly seen in LLMs. As a result, although the final F1-score is slightly lower than dictionary-based systems evaluated on structured job advertisements, the proposed approach provides a more realistic and robust solution for extracting soft skills from heterogeneous CV narratives. These findings validate the effectiveness of the proposed LLM-based pipeline and justify the use of the final optimized configuration for the subsequent analysis of the full CV dataset.

\section{Results}
\label{sec:results}

This section presents the results obtained from applying the optimized soft skill extraction pipeline to the balanced dataset of 300 professional CVs. The analysis relies on the final configuration of the Gemini 3 Pro Preview model, which demonstrated the highest performance during the iterative evaluation process described in the previous section. Each RQ subsection first characterizes the descriptive pattern and then reports the confirmatory tests associated with it; Section~\ref{sec:hyptests} consolidates all hypothesis outcomes and the exploratory contrasts, and Section~\ref{sec:robustness} reports the robustness analyses.

\begin{figure*}
\centerline{\includegraphics[width=0.75\textwidth]{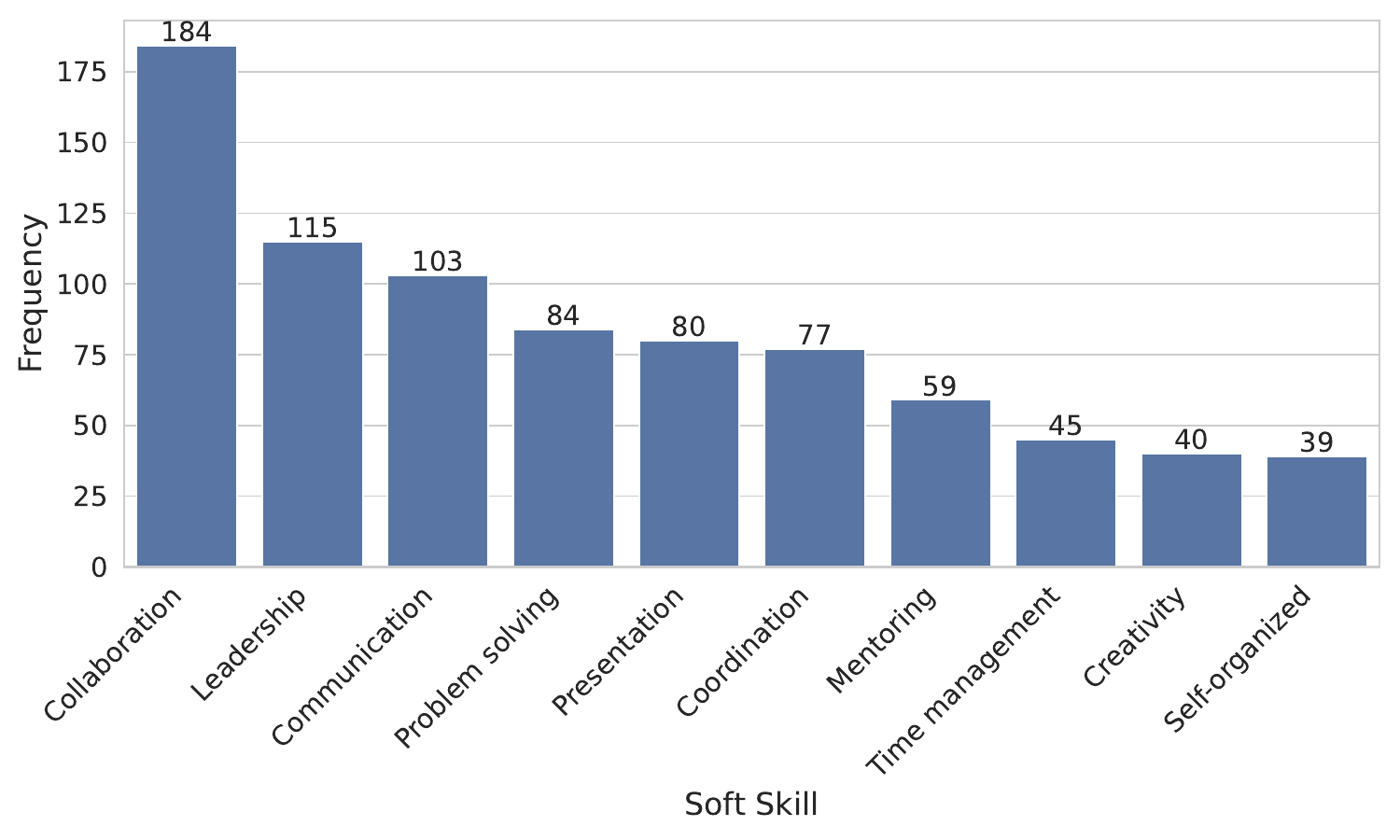}}
\caption{Top 10 soft skills mentioned across all roles}
\label{fig3}
\end{figure*}

\begin{figure*}
\centerline{\includegraphics[width=1\textwidth]{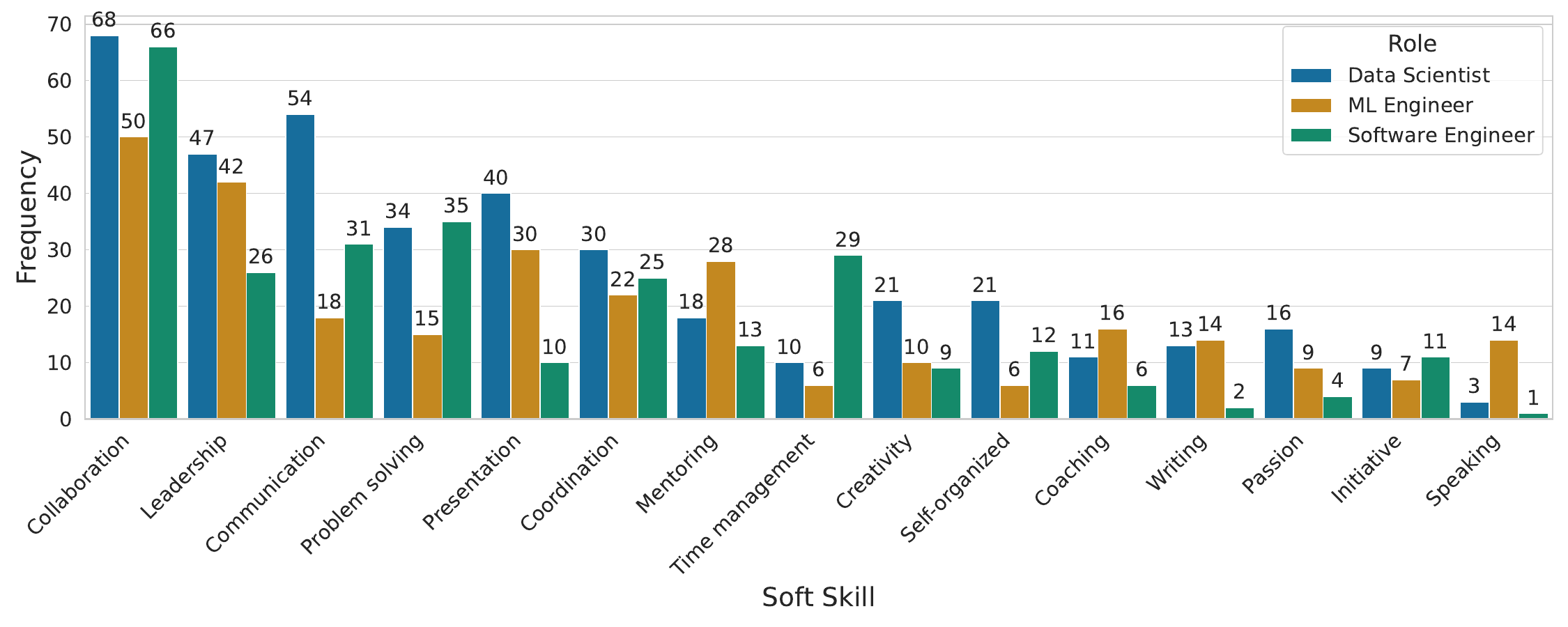}}
\caption{The number of soft skills mentioned by each role}
\label{fig4}
\end{figure*}

\subsection{Soft Skills Across Roles (RQ1)}

Before examining which specific competencies are most prevalent, we first characterize the overall density of soft skill articulation in the dataset. Across the 300 CVs, the optimized extraction pipeline identified 1,367 soft skill mentions in total, corresponding to 1,354 unique skill--CV pairs, an average of approximately 4.5 distinct competencies per CV across the full corpus, or 5.2 when restricting to the 262 CVs (87.3\%, 95\% CI [83.1, 90.6]) that contain at least one detectable soft skill. The remaining 38 CVs (12.7\% [9.4, 16.9]) contained no extractable soft skills at all; this no-skill group is concentrated in two roles, ML engineers (18.0\%) and software engineers (16.0\%), with data scientists exhibiting only 4.0\% no-skill CVs.

The density of soft skill articulation differs notably across the three roles, and the difference is larger than sampling variation would produce. Data scientist CVs are the densest, with a mean of 5.74 distinct competencies per CV ($SD=3.79$), followed by software engineers ($M=4.07$, $SD=3.11$) and ML engineers ($M=3.73$, $SD=3.15$); a Kruskal--Wallis test on the per-CV counts rejects equality across roles, $H(2)=17.39$, $p<0.001$, $\epsilon^2=0.058$. Data scientist CVs also have the longest right tail of skill mentions and the lowest no-skill rate. ML engineering CVs are the sparsest in soft skill content, both because of their high no-skill rate and because the most frequent soft skills among ML engineers (collaboration at 50\%, leadership at 42\%) appear at lower rates than the corresponding figures for data scientists (collaboration 68\%, communication 54\%). The two plausible interpretations of this pattern are that the conservative extraction pipeline requires explicit textual evidence for each skill and that ML engineers more often foreground technical tools and frameworks at the expense of behavioral narrative; Section~\ref{sec:robustness} shows that the first of these cannot account for the observed differences.

To address RQ1, the extracted skills were aggregated across the entire corpus of 300 CVs. The resulting frequency distribution highlights the most prevalent behavioral competencies reported by professionals in the dataset, as shown in Fig.~\ref{fig3}. Collaboration is the most commonly reported soft skill, appearing in 61.3\% of the analyzed CVs (184/300, 95\% CI [55.7, 66.7]). This finding suggests that collaborative capabilities remain a central expectation within modern AI and software development environments, where projects are typically conducted in multidisciplinary teams. The second most frequent skill is leadership, which appears in 38.3\% of CVs (115/300, [33.0, 43.9]), followed by communication, reported in 34.3\% of the documents (103/300, [29.2, 39.9]).

Although the most frequently reported skills appear across all roles, the comparative analysis reveals several role-specific variations in the prominence of secondary competencies. These differences, which are depicted in Fig.~\ref{fig4}, reflect the distinct responsibilities and collaborative contexts associated with each profession, and four of them were predicted in advance.

\textbf{H\textsubscript{1a} (communication).} Communication is reported substantially more often in data science CVs, appearing in 54.0\% of profiles, against 18.0\% of ML engineering and 31.0\% of software engineering CVs. Pooling the two comparison roles, the contrast is 54.0\% versus 24.5\%, a risk difference of $+29.5$ percentage points (95\% CI [$+17.8$, $+40.4$]), OR (odds ratio) $=3.60$ ([2.11, 6.21]), Cohen's $h=0.62$, Fisher's exact $p<0.001$ (Holm-adjusted $p<0.001$). H\textsubscript{1a} is supported, and the effect is the largest role difference we observe. It survives adjustment for seniority (adjusted OR $2.59$ [1.45, 4.64] relative to software engineers, $p=0.001$), while ML engineers fall significantly below software engineers on the same outcome (adjusted OR $0.51$ [0.26, 0.99], $p=0.047$).

\textbf{H\textsubscript{1b} (mentoring).} Mentoring is emphasized more frequently in ML engineering profiles, where it appears in 28.0\% of CVs, compared with 18.0\% for data scientists and 13.0\% for software engineers. Against the pooled comparison group the contrast is 28.0\% versus 15.5\%, a risk difference of $+12.5$ points ([$+2.8$, $+22.9$]), OR $=2.11$ ([1.13, 3.94]), $h=0.31$, $p=0.013$ (Holm-adjusted $p=0.029$). H\textsubscript{1b} is supported, though with a considerably smaller effect than H\textsubscript{1a}, and the confidence interval admits differences as small as three percentage points.

\textbf{H\textsubscript{1c} (no-skill rate).} The distribution of CVs without identifiable soft skills varies across the three roles: 18 ML engineering CVs (18.0\%), 16 software engineering CVs (16.0\%), and only 4 data science CVs (4.0\%). The omnibus test rejects homogeneity, $\chi^2(2)=10.37$, $p=0.006$ (Holm-adjusted $p=0.029$), Cramér's $V=0.19$. The planned contrast confirms the predicted direction: data scientists are far less likely than the other two roles combined to produce a CV with no detectable behavioral content (4.0\% vs.\ 17.0\%, OR $=0.20$ [0.05, 0.60], risk difference $-13.0$ points [$-19.3$, $-5.6$], $p<0.001$). H\textsubscript{1c} is supported. The effect persists when seniority is held constant (adjusted OR $0.22$ [0.07, 0.69], $p=0.010$), whereas ML engineers are statistically indistinguishable from software engineers once seniority is controlled (adjusted OR $1.03$ [0.48, 2.20], $p=0.94$), indicating that the apparent ML--software difference in the raw rates is attributable to the larger junior share among ML engineers rather than to the role itself.

\textbf{H\textsubscript{1d} (problem solving).} In the dataset, 34.0\% of data science CVs describe problem solving as a behavioral competency, whereas the skill appears in only 15.0\% of ML engineering CVs: a risk difference of $+19.0$ points ([$+7.1$, $+30.3$]), OR $=2.90$ ([1.40, 6.25]), $h=0.45$, $p=0.003$ (Holm-adjusted $p=0.020$). H\textsubscript{1d} is supported. This disparity suggests that data scientists are more likely to present analytical reasoning and experimentation as behavioral strengths, emphasizing their ability to approach complex data challenges through structured problem-solving processes.

\textbf{H\textsubscript{1e} (leadership: predicted invariance) is refuted.} We had predicted, following job-advertisement and interview studies that treat leadership as a universal expectation~\cite{Galster_2022, Malinen_2025}, that leadership prevalence would not differ across roles. It does. Leadership appears in 47.0\% of data science CVs, 42.0\% of ML engineering CVs, and 26.0\% of software engineering CVs; the omnibus test rejects homogeneity, $\chi^2(2)=10.18$, $p=0.006$ (Holm-adjusted $p=0.029$), $V=0.18$. Pairwise contrasts (Holm-corrected within the three) locate the difference. Data scientists exceed software engineers (OR $=2.51$ [1.34, 4.79], $p_{\text{Holm}}=0.010$) and ML engineers exceed software engineers (OR $=2.05$ [1.09, 3.93], $p_{\text{Holm}}=0.0496$), while data scientists and ML engineers do not differ (OR $=0.82$ [0.45, 1.48], $p=0.57$). The equivalence tests agree with this reading: at a $\pm$20-point margin, only the ML--data-science pair is statistically equivalent ($p_{\text{TOST}}=0.016$), whereas the two contrasts involving software engineers are not ($p_{\text{TOST}}=0.27$ and $0.56$). The role difference also survives adjustment for seniority, with both data scientists and ML engineers showing roughly two-and-a-half times the odds of software engineers (adjusted ORs $2.55$ [1.38, 4.69] and $2.41$ [1.29, 4.49]). Leadership is therefore consistently prominent in the two ML-adjacent roles but markedly less so in software engineering CVs, and the demand-side assumption of universality does not transfer to candidate self-presentation.

\begin{tcolorbox}[title=Key takeaways for RQ1, colback=gray!5!white, colframe=black!50!white]
\begin{itemize}
\item Collaboration, leadership, and communication dominate across all roles.
\item Communication and problem solving are markedly more prominent in data scientist CVs than in the other two roles (H\textsubscript{1a}, H\textsubscript{1d} supported), and mentoring is more prominent in ML engineer CVs (H\textsubscript{1b} supported).
\item ML engineer CVs are the most \enquote{soft-skill-silent,} and data scientist CVs the least (H\textsubscript{1c} supported); the ML--software difference is explained by seniority composition rather than by role.
\item Contrary to prediction, leadership articulation is not role-invariant (H\textsubscript{1e} refuted): software engineers report it at roughly half the rate of the other two roles.
\end{itemize}
\end{tcolorbox}

\begin{table*}[t]
\centering
\caption{Top 10 Soft Skills by Seniority Level. In the senior column, self-organization ties strategic thinking at rank 10 (26 CVs, 14.0\%); strategic thinking is listed because it is the subject of a confirmatory hypothesis.}
\label{tab4}
\begin{tabular}{lcc|lcc}
\hline
\multicolumn{3}{c|}{\textbf{Junior Professionals (N=114)}} & 
\multicolumn{3}{c}{\textbf{Senior Professionals (N=186)}} \\
\hline
\textbf{Soft Skill} & \textbf{Count} & \textbf{Percentage} & 
\textbf{Soft Skill} & \textbf{Count} & \textbf{Percentage} \\
\hline
Collaboration & 58 & 50.9\% & Collaboration & 126 & 67.7\% \\
Communication & 30 & 26.3\% & Leadership & 88 & 47.3\% \\
Leadership & 27 & 23.7\% & Communication & 73 & 39.2\% \\
Presentation & 24 & 21.1\% & Problem solving & 61 & 32.8\% \\
Problem solving & 23 & 20.2\% & Coordination & 58 & 31.2\% \\
Coordination & 19 & 16.7\% & Presentation & 56 & 30.1\% \\
Mentoring & 17 & 14.9\% & Mentoring & 42 & 22.6\% \\
Time management & 16 & 14.0\% & Creativity & 29 & 15.6\% \\
Adaptive & 15 & 13.2\% & Time management & 29 & 15.6\% \\
Motivated & 15 & 13.2\% & Strategic thinking & 26 & 14.0\% \\
\hline
\end{tabular}
\end{table*}

\subsection{Variation by Seniority (RQ2)}

To answer RQ2, candidates were categorized using a five-year experience threshold. Professionals with less than five years of experience were classified as junior, whereas those with five or more years were categorized as senior. The resulting split is 114 junior and 186 senior CVs. Within the software engineering cohort, 36 profiles were classified as junior and 64 as senior. In the data science group, 32 candidates were junior, and 68 were senior. The ML engineering cohort exhibits the highest proportion of junior profiles, with 46 juniors and 54 seniors. This comparatively larger share of early-career professionals may indicate a relatively recent expansion of entry-level opportunities in the ML domain.

Senior CVs are denser in behavioral content overall. Senior professionals report a mean of 4.99 distinct competencies ($SD=3.48$) against 3.74 ($SD=3.31$) for junior professionals, Mann--Whitney $U=12{,}971$, $p=0.001$, rank-biserial $r=0.22$. Table~\ref{tab4} summarizes the 10 most frequently mentioned soft skills across junior and senior cohorts for the entire dataset. Four of the five directional predictions concern specific competencies within this shift.

\textbf{H\textsubscript{2a} (leadership) is supported, with the largest seniority effect in the study.} Aggregated across roles, leadership appears in 23.7\% of junior CVs (27/114) and 47.3\% of senior CVs (88/186): a risk difference of $+23.6$ points (95\% CI [$+12.5$, $+33.5$]), OR $=2.88$ ([1.68, 5.06]), $h=0.50$, Fisher's exact $p<0.001$ (Holm-adjusted $p<0.001$). This shift indicates that as individuals gain experience, their CVs evolve from documenting individual technical contributions toward demonstrating team guidance, project coordination, and strategic impact.

\textbf{H\textsubscript{2b} (coordination) and H\textsubscript{2c} (strategic thinking) are supported.} Coordination rises from 16.7\% to 31.2\% (OR $=2.26$ [1.23, 4.30], risk difference $+14.5$ points [$+4.5$, $+23.5$], $h=0.34$, $p=0.006$, Holm-adjusted $p=0.029$), and strategic thinking from 5.3\% to 14.0\% (OR $=2.92$ [1.12, 8.96], risk difference $+8.7$ points [$+1.6$, $+15.1$], $h=0.30$, $p=0.020$, Holm-adjusted $p=0.029$). Strategic thinking is the rarest of the confirmatory outcomes, and its wide odds-ratio interval reflects that; the direction is nonetheless the predicted one, and the interval excludes unity.

\textbf{H\textsubscript{2d} (no-skill rate) is supported.} Among senior professionals, only 7.5\% (14/186) of CVs lacked detectable behavioral competencies, against 21.1\% (24/114) of junior CVs: OR $=0.31$ ([0.14, 0.65]), risk difference $-13.5$ points ([$-22.4$, $-5.5$]), $h=-0.40$, $p=0.001$ (Holm-adjusted $p=0.009$). The discrepancy is particularly evident among junior ML engineers, where the no-skill rate reaches 30.4\% against 7.4\% for senior ML engineers.

\textbf{H\textsubscript{2e} adjudicates between substitution and accumulation, and accumulation wins.} Collaboration does not recede as leadership language grows; it increases alongside it, from 50.9\% among junior professionals to 67.7\% among senior professionals (OR $=2.02$ [1.22, 3.37], risk difference $+16.9$ points [$+5.5$, $+27.9$], $h=0.35$, $p=0.005$, Holm-adjusted $p=0.029$). The substitution account, which would have required a negative difference, is therefore refuted. Senior CVs add directional language to collaborative language rather than replacing one with the other. This has a direct practical implication, discussed in Section~\ref{sec:discussion}, for what junior candidates should infer about how to write about collaboration.

\textbf{H\textsubscript{2f} (the seniority effect is not an artifact of role composition) is supported.} Because the three roles differ both in leadership prevalence (H\textsubscript{1e}) and in seniority composition, the aggregate seniority effect could in principle be a compositional artifact. It is not. In a logistic model with role held constant, seniority remains a strong predictor of leadership articulation, adjusted OR $=3.06$ (95\% CI [1.80, 5.23]), likelihood-ratio $\chi^2(1)=18.21$, $p<0.001$ (Holm-adjusted $p<0.001$). The non-parametric analogue agrees: the Cochran--Mantel--Haenszel common odds ratio stratified by role is $3.04$ ([1.79, 5.17]), $\chi^2=17.40$, $p<0.001$.

The second component of H\textsubscript{2f} concerns homogeneity, and here the inferential analysis corrects an impression that the descriptive statistics invite. Read as raw percentages, the seniority transition looks most dramatic among ML engineers, where leadership mentions rise from 23.9\% (11/46) to 57.4\% (31/54), against 34.4\% to 52.9\% for data scientists and 13.9\% to 32.8\% for software engineers. The corresponding stratum-specific odds ratios are $4.22$ ([1.67, 11.30]) for ML engineers, $3.00$ ([0.96, 11.30]) for software engineers, and $2.13$ ([0.83, 5.71]) for data scientists. These intervals overlap heavily, and both formal tests of homogeneity fail to reject. The seniority~$\times$~role interaction contributes nothing to the model (likelihood-ratio $\chi^2(2)=1.22$, $p=0.54$), and the Breslow--Day test gives $\chi^2(2)=1.22$, $p=0.54$. The data are therefore consistent with a single, common seniority effect operating in all three roles; the apparent concentration among ML engineers is a consequence of their lower junior baseline, not evidence of a steeper trajectory. We flag this explicitly because the descriptive reading is the one a frequency-only analysis would have produced, and it would have been wrong.

\begin{tcolorbox}[title=Key takeaways for RQ2, colback=gray!5!white, colframe=black!50!white]
\begin{itemize}
\item Leadership mentions roughly double from junior to senior CVs (23.7\% $\rightarrow$ 47.3\%; OR $2.88$), and the effect survives adjustment for role (adjusted OR $3.06$).
\item Coordination, strategic thinking, and the disappearance of soft-skill-silent CVs follow the same direction (H\textsubscript{2b}--H\textsubscript{2d} supported).
\item Collaboration accumulates rather than being displaced by leadership (H\textsubscript{2e}): senior CVs report both more often.
\item The seniority effect is statistically homogeneous across roles; the apparently steeper ML engineering trajectory reflects a lower junior baseline, not a different slope.
\end{itemize}
\end{tcolorbox}

\subsection{Presentation Style (RQ3)}

Concerning RQ3, the extracted skills were analyzed according to their presentation style within the CV. As described in the methodology, the extraction pipeline differentiates between explicit mentions, where soft skills appear in dedicated sections such as \enquote{Skills}, and implicit demonstrations, where behavioral competencies are conveyed through professional narratives in sections such as work experience or project descriptions. This distinction enables the identification of how technical professionals communicate their behavioral capabilities and whether they rely more on structured keyword lists or contextual storytelling.

\begin{figure*}
\centerline{\includegraphics[width=0.7\textwidth]{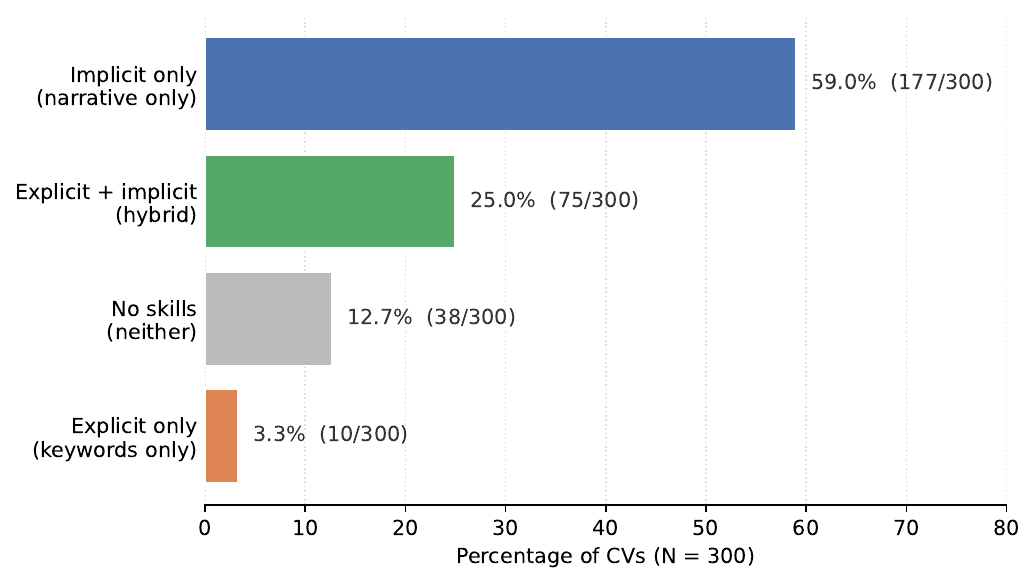}}
\caption{Distribution of soft skill disclosure styles across the CVs.}
\label{fig5}
\end{figure*}

The results reveal a strong preference for narrative-based disclosure of soft skills. Across the dataset of 300 CVs, candidates were nearly three times more likely to demonstrate soft skills through professional narratives than to explicitly list them as standalone keywords. Overall, only 3.3\% of candidates (10/300) relied exclusively on explicit keyword lists as illustrated in Fig.~\ref{fig5}. In contrast, 59.0\% of the CVs (177/300) communicated behavioral competencies entirely through contextual descriptions of their professional experiences without listing them in dedicated skills sections. When examining each disclosure style in more detail, explicit mentions were identified in 28.3\% of CVs (85/300). The most frequently listed skills within these structured sections were communication (15.7\%) and collaboration (13.7\%), indicating that candidates occasionally highlight widely recognized interpersonal competencies as explicit labels. However, implicit demonstrations were considerably more prevalent, appearing in 84.0\% of CVs (252/300). The skills most frequently conveyed through narrative descriptions were collaboration (47.7\%), leadership (33.7\%), and coordination (24.7\%), typically illustrated through statements describing collaboration on projects, team leadership responsibilities, or cross-functional coordination. Additionally, 25.0\% of candidates (75/300) adopted a hybrid disclosure strategy, combining explicit skill listings with supporting narrative evidence in their professional experience sections.

\textbf{H\textsubscript{3a} is supported by a very large margin.} Because explicit and implicit disclosure are recorded on the same document, the comparison is paired. Of the 187 CVs whose two channels disagree, 177 disclose behavioral competencies only through narrative and 10 only through keyword lists, a discordant share of 94.7\% (95\% CI [90.4, 97.1]); the exact McNemar test gives $p<0.001$ with an odds ratio of $17.7$. The same conclusion holds when the comparison is made on counts rather than presence. The per-CV number of narratively disclosed competencies exceeds the number of keyword-disclosed competencies in the large majority of the 250 CVs where the two differ, Wilcoxon signed-rank $p<0.001$, $r=0.60$.

This aggregate result conceals an asymmetry that matters more for practice than the aggregate itself. Disaggregating by competency (Table~\ref{tab5}) shows that narrative dominance is not uniform across the taxonomy but is concentrated in precisely those competencies employers describe as decisive. Coaching (100\% narrative), coordination (96.1\%), mentoring (94.9\%), presentation (90.0\%), and leadership (87.8\%) are disclosed almost exclusively through narrative, and all remain significant after false-discovery correction. At the other end, communication (54.4\% narrative, $q=0.54$), problem-solving (59.5\%, $q=0.15$), adaptability (52.8\%), self-organization (51.3\%), and analytical ability (34.4\% narrative, i.e., predominantly keyword-listed) are as likely or more likely to be stated as labels. In other words, the competencies that candidates put in a skills section are largely the generic ones, while the specific, evidence-bearing competencies that distinguish a senior candidate are the ones that only appear in prose.

\begin{figure*}
\centerline{\includegraphics[width=0.9\textwidth]{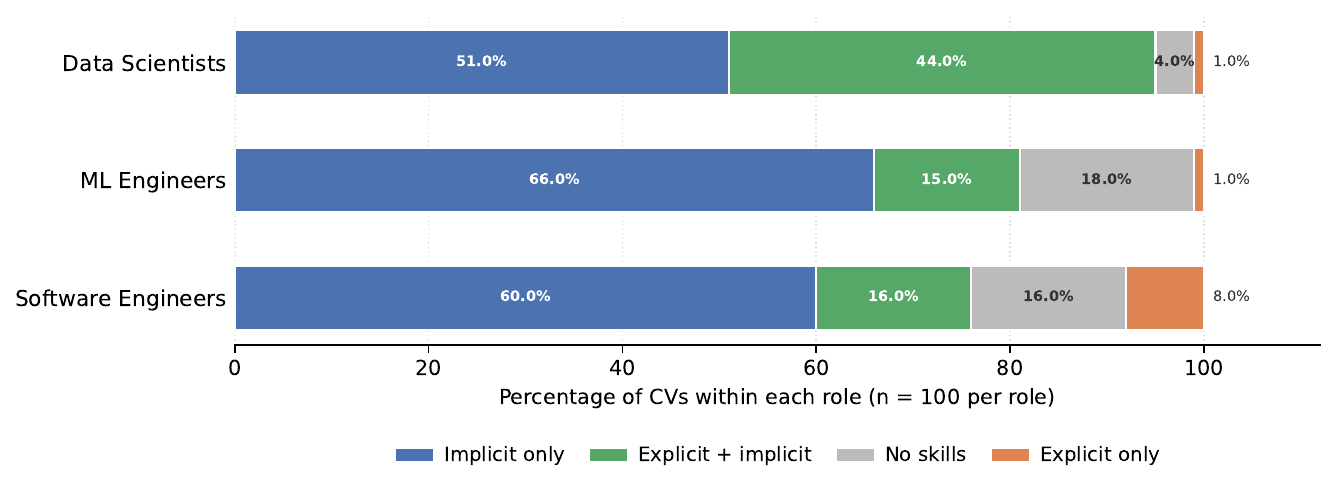}}
\caption{Soft skill disclosure styles by role.}
\label{fig6}
\end{figure*}

\textbf{H\textsubscript{3b} is partially supported.} The keyword--narrative ratio varies across the three analyzed roles, as shown in Fig.~\ref{fig6}, and the $3\times4$ test of disclosure style by role rejects independence, $\chi^2(6)=42.46$, $p<0.001$ (Holm-adjusted $p<0.001$), $V=0.27$. Disclosure style, therefore, does depend on role. The second clause of the hypothesis, that ML engineers rely on narrative most heavily, is not confirmed at conventional levels. ML engineers show the highest narrative-only rate, with 66.0\% of CVs demonstrating soft skills solely through descriptions of professional activities, against 60.0\% for software engineers and 51.0\% for data scientists, but neither the omnibus contrast on narrative-only rates ($\chi^2(2)=4.71$, $p=0.095$) nor the planned contrast of ML engineers against the other two roles (66.0\% vs.\ 55.5\%, OR $=1.55$ [0.92, 2.66], $p=0.083$) reaches significance. What the significant $3\times4$ result reflects is a different contrast: data scientists are far more likely to use both channels (44.0\%, against 15.0\% for ML engineers and 16.0\% for software engineers) and far less likely to use neither. These candidates frequently label competencies such as communication directly and then reinforce them through examples of presenting analytical findings or collaborating with stakeholders. Software engineers, meanwhile, show a slightly higher tendency toward structured keyword presentation, with 8.0\% of CVs relying exclusively on explicit skill lists, against 1.0\% in each of the other two roles. Only 16.0\% of ML engineering CVs list any behavioral competency in a dedicated skills section.

Disclosure style does not vary significantly with seniority. Narrative-only disclosure rises from 53.5\% among junior professionals to 62.4\% among senior professionals, but the difference is not significant ($p=0.15$), and the proportion of CVs containing any explicit skills section is essentially flat (25.4\% junior vs.\ 30.1\% senior, $p=0.43$). What changes with seniority is the amount and kind of behavioral content, not the channel through which it is delivered.

Overall, these findings indicate that the communication of soft skills in technical CVs relies predominantly on contextual storytelling rather than simple keyword enumeration. As professionals accumulate experience, their CVs shift from listing basic collaborative traits toward narrating more complex responsibilities, such as leadership, coordination, and project orchestration. Furthermore, the substantial number of implicitly disclosed competencies in the corpus (1,007 skill--CV pairs) compared with explicitly listed ones (347) indicates that roughly three-quarters (74.4\%) of the soft skill evidence in these CVs takes a form that keyword matching cannot detect. This is a property of the documents themselves rather than of any particular tool. Any screening approach restricted to explicit labels observes only a small fraction of the competencies candidates actually convey, and, as Table~\ref{tab5} shows, the fraction it does observe is biased toward the least discriminating competencies.

\begin{tcolorbox}[title=Key takeaways for RQ3, colback=gray!5!white, colframe=black!50!white]
\begin{itemize}
\item Soft skills are predominantly conveyed through narrative storytelling, not keyword lists (H\textsubscript{3a}; 94.7\% of discordant CVs, exact McNemar $p<0.001$).
\item Narrative dominance is concentrated in the highest-value competencies: leadership, coordination, mentoring, and coaching are 88--100\% narrative, whereas communication, problem solving, and analytical ability are as often keyword-listed.
\item Disclosure style depends on role (H\textsubscript{3b}, first clause), but the driver is the data scientists' dual-channel style rather than a significant ML engineering excess in narrative-only disclosure (second clause not confirmed).
\item Disclosure style does not change significantly with seniority; the volume and kind of behavioral content do.
\end{itemize}
\end{tcolorbox}

\begin{table*}[t]
\centering
\caption{Disclosure channel by competency. \enquote{Keyword} and \enquote{Narrative} partition the CVs in which the competency was detected, so their sum is the competency's total prevalence. \enquote{Narrative share} is the percentage of those CVs disclosing it through narrative rather than a dedicated skills section, with a 95\% Wilson interval. $p$ is the two-sided exact binomial test against a 50/50 split; $q$ is the Benjamini--Hochberg adjusted value. Competencies detected in fewer than 15 CVs are omitted.}
\label{tab5}
\small
\begin{tabular}{|l|r|r|r|l|r|r|}
\hline
\textbf{Competency} & \textbf{Keyword} & \textbf{Narrative} & \textbf{Narrative share} & \textbf{95\% CI} & \textbf{$p$} & \textbf{$q$} \\
\hline
Coaching & 0 & 33 & 100.0\% & [89.6, 100.0] & $<0.001$ & $<0.001$ \\
\hline
Coordination & 3 & 74 & 96.1\% & [89.2, 98.7] & $<0.001$ & $<0.001$ \\
\hline
Mentoring & 3 & 56 & 94.9\% & [86.1, 98.3] & $<0.001$ & $<0.001$ \\
\hline
Accountability & 1 & 15 & 93.8\% & [71.7, 98.9] & $<0.001$ & 0.001 \\
\hline
Presentation & 8 & 72 & 90.0\% & [81.5, 94.8] & $<0.001$ & $<0.001$ \\
\hline
Initiative & 3 & 24 & 88.9\% & [71.9, 96.1] & $<0.001$ & $<0.001$ \\
\hline
Speaking & 2 & 16 & 88.9\% & [67.2, 96.9] & 0.001 & 0.002 \\
\hline
Leadership & 14 & 101 & 87.8\% & [80.6, 92.6] & $<0.001$ & $<0.001$ \\
\hline
Writing & 4 & 25 & 86.2\% & [69.4, 94.5] & $<0.001$ & $<0.001$ \\
\hline
Strategic thinking & 6 & 26 & 81.3\% & [64.7, 91.1] & $<0.001$ & 0.001 \\
\hline
Time management & 9 & 36 & 80.0\% & [66.2, 89.1] & $<0.001$ & $<0.001$ \\
\hline
Creativity & 8 & 32 & 80.0\% & [65.2, 89.5] & $<0.001$ & $<0.001$ \\
\hline
Goal orientation & 5 & 18 & 78.3\% & [58.1, 90.3] & 0.011 & 0.018 \\
\hline
Collaboration & 41 & 143 & 77.7\% & [71.2, 83.1] & $<0.001$ & $<0.001$ \\
\hline
Curiosity & 9 & 23 & 71.9\% & [54.6, 84.4] & 0.020 & 0.032 \\
\hline
Passion & 11 & 18 & 62.1\% & [44.0, 77.3] & 0.265 & 0.353 \\
\hline
Problem solving & 34 & 50 & 59.5\% & [48.8, 69.4] & 0.101 & 0.152 \\
\hline
Communication & 47 & 56 & 54.4\% & [44.8, 63.7] & 0.431 & 0.544 \\
\hline
Adaptive & 17 & 19 & 52.8\% & [37.0, 68.0] & 0.868 & 0.992 \\
\hline
Self-organized & 19 & 20 & 51.3\% & [36.2, 66.1] & 1.000 & 1.000 \\
\hline
Decision making & 11 & 11 & 50.0\% & [30.7, 69.3] & 1.000 & 1.000 \\
\hline
Motivated & 16 & 15 & 48.4\% & [32.0, 65.2] & 1.000 & 1.000 \\
\hline
Social skills & 10 & 7 & 41.2\% & [21.6, 64.0] & 0.629 & 0.755 \\
\hline
Analytical & 21 & 11 & 34.4\% & [20.4, 51.7] & 0.110 & 0.156 \\
\hline
\end{tabular}
\end{table*}

\subsection{Summary of Hypothesis Tests and Exploratory Contrasts}
\label{sec:hyptests}

Table~\ref{tab:hypresults} consolidates the outcome of every confirmatory test. Eleven of the 13 hypotheses are supported, one (H\textsubscript{3b}) is partially supported, and one (H\textsubscript{1e}) is refuted. All 13 raw $p$-values remain significant after Holm--Bonferroni correction across the family, so the family-wise error control does not change any conclusion; we report the adjusted values nonetheless because the strength of evidence differs substantially across the family, ranging from $p_{\text{Holm}}<10^{-5}$ for H\textsubscript{3a} and H\textsubscript{1a} to $p_{\text{Holm}}=0.029$ for six hypotheses that sit close to the corrected threshold.

Two of the outcomes deserve emphasis because they run against the interpretation that a descriptive analysis would have produced. First, H\textsubscript{1e} was formulated as a prediction of no role difference, taken directly from the demand-side literature, and it is refuted. Leadership articulation is not universal across roles but concentrated in the two ML-adjacent ones. Second, H\textsubscript{2f}'s homogeneity clause shows that the seniority effect on leadership does not differ across roles, which contradicts the natural reading of the raw percentages. A study that reported frequencies alone would plausibly have claimed both that leadership is a universal baseline and that ML engineers show a distinctive seniority trajectory; neither claim survives testing.

\begin{table*}[t]
\centering
\caption{Outcomes of the 13 confirmatory hypothesis tests. Effect sizes are odds ratios (OR) with exact 95\% confidence intervals for $2\times2$ contrasts, Cram\'er's $V$ for omnibus $\chi^2$ tests. $p_{\text{Holm}}$ is corrected across the family of 13. Data Scientist (DS), Software Engineer (SE). Percentage triples are reported in the order ML / DS / SE.}
\label{tab:hypresults}
\footnotesize
\setlength{\tabcolsep}{3pt}
\renewcommand{\arraystretch}{1.15}
\begin{tabular}{|L{0.6cm}|L{3.5cm}|L{2.1cm}|L{4.3cm}|L{1.0cm}|L{1.0cm}|L{1.9cm}|}
\hline
\textbf{ID} & \textbf{Contrast} & \textbf{Observed} & \textbf{Effect size [95\% CI]} & \textbf{$p$} & \textbf{$p_{\text{Holm}}$} & \textbf{Outcome} \\
\hline
H\textsubscript{1a} & Communication: DS vs.\ ML+SE & 54.0\% vs.\ 24.5\% & OR 3.60 [2.11, 6.21] & $<0.001$ & $<0.001$ & Supported \\
\hline
H\textsubscript{1b} & Mentoring: ML vs.\ DS+SE & 28.0\% vs.\ 15.5\% & OR 2.11 [1.13, 3.94] & 0.013 & 0.029 & Supported \\
\hline
H\textsubscript{1c} & No-skill rate across roles & 18 / 4 / 16\% & $V=0.19$; contrast OR 0.20 [0.05, 0.60] & 0.006 & 0.029 & Supported \\
\hline
H\textsubscript{1d} & Problem solving: DS vs.\ ML & 34.0\% vs.\ 15.0\% & OR 2.90 [1.40, 6.25] & 0.003 & 0.020 & Supported \\
\hline
H\textsubscript{1e} & Leadership across roles (predicted equal) & 42 / 47 / 26\% & $V=0.18$; DS vs.\ SE OR 2.51 [1.34, 4.79] & 0.006 & 0.029 & Refuted \\
\hline
H\textsubscript{2a} & Leadership: senior vs.\ junior & 47.3\% vs.\ 23.7\% & OR 2.88 [1.68, 5.06] & $<0.001$ & $<0.001$ & Supported \\
\hline
H\textsubscript{2b} & Coordination: senior vs.\ junior & 31.2\% vs.\ 16.7\% & OR 2.26 [1.23, 4.30] & 0.006 & 0.029 & Supported \\
\hline
H\textsubscript{2c} & Strategic thinking: senior vs.\ junior & 14.0\% vs.\ 5.3\% & OR 2.92 [1.12, 8.96] & 0.020 & 0.029 & Supported \\
\hline
H\textsubscript{2d} & No-skill rate: senior vs.\ junior & 7.5\% vs.\ 21.1\% & OR 0.31 [0.14, 0.65] & 0.001 & 0.009 & Supported \\
\hline
H\textsubscript{2e} & Collaboration: senior vs.\ junior & 67.7\% vs.\ 50.9\% & OR 2.02 [1.22, 3.37] & 0.005 & 0.029 & Accumulation \\
\hline
H\textsubscript{2f} & Leadership $\sim$ seniority $\mid$ role & adjusted & OR 3.06 [1.80, 5.23]; interaction $p=0.54$ & $<0.001$ & $<0.001$ & Supported \\
\hline
H\textsubscript{3a} & Narrative vs.\ keyword within CV & 177 vs.\ 10 discordant & OR 17.7; 94.7\% [90.4, 97.1] & $<0.001$ & $<0.001$ & Supported \\
\hline
H\textsubscript{3b} & Disclosure style $\times$ role & $3\times4$ table & $V=0.27$; ML contrast $p=0.083$ & $<0.001$ & $<0.001$ & Partial \\
\hline
\end{tabular}
\end{table*}

Beyond the confirmatory family, we tested every competency detected in at least 15 CVs (24 labels) against role and against seniority, controlling the false discovery rate at 5\% within each family. Seventeen competencies differ significantly across roles after correction. The strongest are communication ($V=0.31$, $q<0.001$), presentation ($V=0.28$, $q<0.001$), and time management ($V=0.28$, $q<0.001$), the last of these running in the opposite direction to most role effects: it is a software engineering signature (29\% vs.\ 10\% and 6\%). Speaking ($V=0.24$, $q=0.001$) and writing ($V=0.18$, $q=0.016$) are disproportionately ML engineering markers, while strategic thinking ($V=0.23$, $q=0.002$), decision making ($V=0.18$, $q=0.016$), self-organization ($V=0.18$, $q=0.016$), curiosity ($V=0.18$, $q=0.019$), and passion ($V=0.17$, $q=0.028$) cluster in data science CVs. In the seniority family, only leadership survives correction ($q=0.001$); coordination and collaboration fall just outside the threshold ($q=0.0501$ for both), and communication, strategic thinking, and problem solving are suggestive but not significant after adjustment ($q=0.098$ each). We note this asymmetry plainly that the role signature is broad and distributed across many competencies, whereas the seniority signal, when the entire taxonomy is scanned without prior direction, is carried overwhelmingly by leadership alone. The confirmatory results for H\textsubscript{2b}, H\textsubscript{2c}, and H\textsubscript{2e} are not undermined by this, since they were specified in advance and belong to a family of 13 rather than 24, but their exploratory counterparts are a reminder that they are the weaker members of the confirmatory set.

\subsection{Robustness and Sensitivity Analyses}
\label{sec:robustness}

Three analyses probe whether the confirmatory conclusions depend on choices made in the measurement pipeline. We focus them on H\textsubscript{2a}, the primary hypothesis, and on H\textsubscript{1a}, the largest role effect.

\textbf{Dichotomization of seniority.} The five-year threshold discards information. Re-estimating the primary hypothesis with the underlying continuous estimate of total experience, in a logistic model adjusting for role, reproduces the effect. Each additional year of experience is associated with a 22\% increase in the odds of articulating leadership, OR $=1.22$ per year (95\% CI [1.13, 1.33]), $p<0.001$, $n=295$ (five CVs lack a usable experience estimate). The seniority result is therefore not an artifact of where the threshold was placed.

\textbf{Document length.} Longer CVs offer more opportunities for a competency to be detected, and senior CVs are longer on average. Adding document length to the model leaves the seniority effect essentially unchanged (adjusted OR $2.82$ [1.60, 4.97], $p<0.001$), and length itself is not a significant predictor once seniority and role are in the model (OR $1.02$ per 1,000 characters, $p=0.41$).

\textbf{Extraction error.} The most substantive threat is that the outcome variable is produced by a pipeline with a measured recall of 0.74 and precision of 0.69. We address it in two steps rather than by argument alone.

First, we propagate the measurement error through the estimates. Treating extraction as a diagnostic test applied to each competency--CV cell, the measured precision and recall together with the observed cell-level prevalence imply a specificity of approximately 0.967 (the value ranges from 0.958 to 0.973 across plausible assumptions about the effective size of the taxonomy). Applying the standard matrix correction for non-differential misclassification, $\pi = (p_{\text{obs}} + \mathit{Sp} - 1)/(\mathit{Se} + \mathit{Sp} - 1)$, every confirmatory contrast becomes larger, not smaller. The leadership seniority gap widens from $+23.6$ to $+33.4$ percentage points, the communication role gap from $+29.5$ to $+41.7$ points, the leadership gap between data scientists and software engineers from $+21.0$ to $+29.7$ points, and the collaboration seniority gap from $+16.9$ to $+23.9$ points. Varying sensitivity over $[0.65, 0.85]$ and specificity over $[0.95, 0.99]$ places the corrected leadership seniority gap between $+28.1$ and $+39.4$ points, in every case above the uncorrected estimate. Imperfect extraction attenuates the observed differences toward zero, so the reported effects are conservative rather than inflated.

Second, non-differential error is the favorable case; the real concern is differential error, in which the pipeline detects competencies more readily in one group than another. We therefore compute how large such a differential would have to be to eliminate each effect. For H\textsubscript{2a}, no differential favoring senior CVs can do it; even a pipeline with perfect sensitivity on senior CVs and the measured sensitivity of 0.74 on junior CVs would leave a corrected gap of 16.7 percentage points, and closing the gap entirely would require a sensitivity above 1.0, which is impossible. Running the correction in the other direction, sensitivity on junior CVs would have to fall to 0.36, less than half the value measured on the evaluation set, while remaining 0.74 on senior CVs. For H\textsubscript{1a} the corresponding figures are 1.72 (impossible) and 0.33. Differential extraction error of that magnitude is not credible for a pipeline whose prompt is identical across groups and whose evaluation sample spans all three roles and both seniority levels.

Two limits of this analysis should be stated. It assumes that the pipeline's aggregate sensitivity and specificity, measured over the taxonomy as a whole, apply to each competency individually, which will not be exactly true for rare labels; and it corrects prevalences, not the tests themselves, so the corrected risk differences should be read as indicating the direction and rough magnitude of the bias rather than as replacement point estimates. Neither limit affects the qualitative conclusion, which is that the reported effects are attenuated by extraction noise rather than manufactured by it. Finally, the explicit-versus-implicit comparison in RQ3 is internally consistent in a way that makes it especially robust. Both branches rely on the same extraction model and the same taxonomy, so systematic extraction bias affects both sides of the comparison and largely cancels in the implicit-to-explicit ratio.

\section{Discussion}
\label{sec:discussion}

The findings of this study provide several insights into how soft skills are articulated in technical CVs across ML-related professions. First, the aggregated analysis confirms that collaboration, leadership, and communication represent the most prominent behavioral competencies across the dataset. The strong presence of collaboration-related skills supports the view that modern AI and software engineering projects rely heavily on multidisciplinary collaboration, where ML engineers, data scientists, and software engineers must work closely to deliver complex systems. Leadership and communication also appear frequently, reinforcing prior research suggesting that interpersonal competencies are increasingly valued alongside technical expertise in engineering roles.

\subsection{Role signatures and where the demand-side account holds}

Three of the four directional role predictions we derived from prior work are borne out on the candidate side, and with effects large enough to be practically meaningful. Data science profiles place greater emphasis on communication (OR 3.60) and problem solving (OR 2.90), reflecting the need to interpret analytical findings and communicate insights to non-technical stakeholders. ML engineering CVs highlight mentoring more frequently (OR 2.11), which may reflect the bridging role that ML engineers often play between software engineering and data science teams. That these candidate-side patterns match what the job-advertisement and interview literature attribute to the same roles is a non-trivial corroboration. The two data sources are collected independently, from different populations, using different instruments, and there was no guarantee that employer-side role descriptions would leave a signature in how individuals write about themselves.

The one prediction that fails is instructive precisely because it fails. H\textsubscript{1e} held that leadership would be role-invariant, on the strength of studies that report leadership as a universal expectation in software and AI work~\cite{Galster_2022, Malinen_2025}. In our corpus, it is not: software engineers articulate leadership at roughly half the rate of data scientists and ML engineers (26.0\% vs.\ 47.0\% and 42.0\%), and the gap persists when seniority is held constant. Two readings are available. On the first, the demand-side observation remains correct and the divergence is one of self-presentation. Software engineers may be as likely to exercise leadership but less likely to frame it as such in a document that competes for space with technical depth. On the second, the universality claim was always an aggregation artifact of job-advertisement corpora, in which the leadership language of senior and managerial postings is averaged over a role population that is much broader than the individual-contributor software engineers our sampling captured. Our data cannot separate these, but either reading has the same consequence for practice. A software engineering candidate who leads work and does not say so is systematically less visible than a data science candidate doing the same amount of leading. This is the kind of claim that only candidate-side data can produce, and only inferential analysis can support.

\subsection{Soft-skill-silent CVs and the articulation gap}

Another important observation concerns the presence of CVs without detectable soft skills. Although the majority of CVs (87.3\%) included at least one behavioral competency, a subset, particularly within ML engineering, focused almost exclusively on technical content. This pattern may reflect differences in self-presentation strategies, with some candidates prioritizing the listing of tools, frameworks, and technical achievements over the description of interpersonal competencies. The relatively higher no-skill rate among ML engineers may also indicate that behavioral traits are sometimes expressed using language that falls outside existing taxonomy variants. The inferential analysis adds a qualification that the raw rates obscure. Once seniority is held constant, ML engineers and software engineers are statistically indistinguishable on this outcome (adjusted OR 1.03, $p=0.94$), and it is data scientists who stand apart (adjusted OR 0.22, $p=0.010$). The elevated ML no-skill rate is thus largely a consequence of that cohort containing the most junior professionals, not of something intrinsic to the ML engineering role.

This pattern has practical implications for ML and MLOps practitioners preparing CVs for the technical labor market. Prior work on hiring manager perspectives indicates that non-technical competencies frequently weigh heavily in evaluation decisions for engineering roles, particularly at the early-career stage~\cite{Loufek_2025}, and analyses of job advertisements consistently show that employers explicitly require interpersonal and behavioral skills alongside technical expertise~\cite{Calanca_2019, Galster_2022}. The 18\% no-skill rate observed among ML engineering CVs in our dataset, and the 30.4\% rate among junior ML engineers in particular, therefore represents a meaningful gap between what hiring managers value and what a substantial subset of candidates choose to surface in their CVs. Practitioners who concentrate their CV content on tools, frameworks, and model architectures while omitting any behavioral narrative may be underselling competencies that recruiters actively look for, and that the candidates themselves likely demonstrate in practice through team collaboration, mentoring, and stakeholder communication. We therefore encourage ML and MLOps practitioners, especially those at earlier career stages, to make space in their CVs for explicit or narrative evidence of how they collaborate, lead, mentor, and communicate, even briefly, alongside the technical content.

\subsection{What changes with seniority, and what does not}

The analysis of seniority levels further highlights the evolution of soft skill narratives throughout a professional career. Junior professionals tend to emphasize collaborative participation, whereas senior professionals increasingly highlight leadership, coordination, and strategic competencies. The effect on leadership is substantial and robust. An adjusted odds ratio of 3.06, equivalent to a 23.6-point rise in prevalence, which survives adjustment for role, replicates when experience is treated as a continuous variable, and would require an implausible degree of differential extraction error to explain away.

The inferential analysis corrects the aggregate picture in two ways that matter. First, we had expected, and the descriptive percentages appear to show, that the seniority transition is sharpest among ML engineers, whose leadership mentions rise from 23.9\% to 57.4\%. The homogeneity tests do not support that reading. The stratum-specific odds ratios are statistically indistinguishable (interaction $p=0.54$; Breslow--Day $p=0.54$), and the apparent steepness among ML engineers follows from their unusually low junior baseline rather than from a different rate of change. The professionalization pattern is best described as a single effect operating across all three roles. Second, the competing accounts of what happens to collaboration are adjudicated cleanly in favor of accumulation. Collaboration does not recede as leadership grows; both increase together, and senior CVs report collaboration at 67.7\% against 50.9\% among juniors. The framing of a career transition \enquote{from participation to direction} is therefore misleading as a description of what candidates write: senior CVs are not less collaborative in their language, they are more of both.

An alternative interpretation of the whole seniority pattern is also worth noting. Rather than developing fundamentally new competencies, senior professionals may primarily develop the ability to articulate competencies they already practiced as juniors. The sharp drop in the no-skill rate from 21.1\% to 7.5\% overall, and from 30.4\% to 7.4\% among ML engineers, even though juniors clearly engage in collaborative, team-based work, supports this articulation-gap interpretation. The competencies were arguably present, but the language for surfacing them was not. The accumulation result strengthens this reading. If seniors were genuinely trading participation for direction, we would expect collaborative language to give way; that it instead grows alongside leadership language is more consistent with seniors having simply become better at describing what they do. This distinction matters because it reframes the seniority transition as partly a CV-writing skill rather than purely a competency-acquisition skill, and CV-writing skill is teachable on a much shorter timescale than five years of experience.

\subsection{Implications for education and curriculum design}

The findings also carry implications for educators and curriculum designers preparing students for AI-related careers. Three patterns in particular warrant attention. First, the strong professionalization gap observed between junior and senior CVs, namely that leadership mentions roughly double from 23.7\% to 47.3\% and the no-skill rate drops from 21.1\% to 7.5\%, suggests that articulating leadership, coordination, and stakeholder communication is something many professionals appear to learn in industry rather than during formal education. Curricula in computer science and data science programs could narrow this gap by integrating explicit instruction on how to identify, develop, and communicate behavioral competencies acquired through coursework, capstone projects, and team-based assignments, rather than treating these as implicit byproducts of technical training~\cite{Azamnouri_2026}. Second, the role-specific differences observed in the dataset suggest that program designers should not treat soft skill development as a single homogeneous topic. Programs targeting different specializations may benefit from emphasizing the specific behavioral competencies that practitioners in those roles most frequently demonstrate, e.g., stakeholder communication and analytical problem framing for data science tracks, and team mentoring and cross-functional coordination for ML engineering and MLOps tracks. The refutation of H\textsubscript{1e} adds a further, less obvious recommendation: because software engineering candidates articulate leadership least often despite leadership being a general expectation in the field, software engineering programs in particular may benefit from treating leadership articulation as an explicit learning outcome rather than assuming it emerges. Third, the high no-skill rate among junior ML engineers (30.4\%) suggests a gap not only in soft skill development but also in CV-writing instruction. Career-services support and professional development modules embedded in technical curricula could equip students with concrete strategies for narrating behavioral competencies in their CVs, including the kind of narrative phrasing identified in Section~\ref{sec:results} as the dominant disclosure style among practitioners. Such interventions would address both the development of behavioral competencies, a long-standing concern in the soft skills literature~\cite{Romanenko_2024}, and the ability to articulate them visibly in CVs, which the present paper identifies as a distinct and previously underexplored dimension of the soft skills gap.

\subsection{Implications for screening pipeline design}

Finally, the results demonstrate that soft skills are predominantly communicated through narrative descriptions rather than explicit keyword lists. The roughly three-to-one ratio between implicitly and explicitly disclosed competencies (1,007 vs.\ 347) is not merely a stylistic observation; it has direct consequences for how CVs are processed in modern hiring pipelines. A substantial fraction of recruitment systems, including applicant tracking systems and CV screening tools, rely heavily on keyword matching against predefined skill vocabularies~\cite{Gugnani_2020, Wosiak_2021}. If candidates predominantly express soft skills through narrative rather than explicit labels, such systems will systematically under-detect interpersonal competencies.

The per-competency breakdown in Table~\ref{tab5} sharpens this concern considerably, and is, in our view, the most actionable finding in the paper. Narrative dominance is not uniform. It is concentrated in exactly the competencies that differentiate candidates. Coaching, coordination, mentoring, presentation, and leadership are disclosed through narrative in 88--100\% of the CVs that disclose them at all, whereas communication, problem-solving, adaptability, self-organization, and analytical ability are as likely, or more likely, to be stated as bare labels. A keyword-based screener, therefore, does not simply see a smaller sample of each candidate's behavioral profile; it sees a systematically distorted one, in which the generic self-descriptors that almost every candidate can claim are visible, and the specific, evidence-bearing competencies that would distinguish a senior candidate from a junior one are invisible. Since leadership is also the competency that changes most with seniority (H\textsubscript{2a}), a keyword-based screener is close to blind on precisely the dimension along which it is being asked to discriminate.

This raises a concrete fairness concern. Candidates whose CV-writing conventions favor storytelling may be disadvantaged relative to candidates who happen to list the same competencies as keywords, even when both groups demonstrate them equally in practice. The effect is not evenly distributed. ML engineers, at 66.0\% narrative-only disclosure and only 16.0\% listing any behavioral competency in a skills section, are most exposed, while data scientists, 44.0\% of whom use both channels, are best protected. We note for accuracy that the role difference in narrative-only rates specifically did not reach significance in our data ($p=0.083$), so this should be read as a pattern that warrants replication rather than an established disparity; the difference in dual-channel use, by contrast, is unambiguous. Generative-LLM-based extraction approaches, as developed in this paper, offer one path toward closing this gap by recovering competencies from narrative context. However, building this capability into production-grade screening tools also requires careful attention to the precision--recall trade-offs and hallucination risks, and to the fairness properties of the extractor itself, since an extractor that recovers narrative evidence unevenly across groups would replace one bias with another.

A methodological observation also follows from these findings. The substantial gap between implicitly (1,007) and explicitly (347) disclosed competencies in our corpus provides direct empirical justification for the choice of LLM-based generative extraction over dictionary or keyword-based approaches. Had we relied on keyword matching alone, we would have captured at most 347 of the 1,354 unique skill--CV pairs in the corpus, missing nearly three-quarters (74.4\%) of the soft skill articulation present in candidate CVs. This is not a hypothetical advantage; it is the difference between concluding that ML engineering CVs are largely silent on soft skills, which a keyword-based system would suggest, and concluding that ML engineering CVs are heavily narrative-driven, which our analysis revealed. More concretely still, a keyword-only analysis would have observed 14 of the 115 leadership disclosures in the corpus and would have had no power whatsoever to detect the seniority effect that constitutes our primary finding. The pipeline thus enables a finding that the methodology of much prior work would not have investigated.

\section{Threats to Validity}
\label{sec:threats-to-validity}

This section outlines the potential limitations of the study and discusses threats to validity to provide a transparent interpretation of the findings.

\subsection{Construct Validity}

A first concern regarding construct validity relates to how soft skills are operationally defined and measured in this study. We rely on the TaxoSoft taxonomy~\cite{Khaouja_2019} as the canonical inventory of soft skill categories. While TaxoSoft is grounded in real labor-market language and has been used as a reference in prior soft-skill extraction research, any taxonomy is inherently constrained by the categories it defines. Behavioral competencies expressed in language that falls outside the taxonomy's variants are systematically excluded from our analysis, which may particularly affect candidates whose self-presentation uses non-standard or domain-specific phrasing.

A second construct concern is the operationalization of seniority through a binary five-year threshold. While this aligns with common industry leveling standards, it abstracts away nuanced differences in expertise within each cohort. We mitigated this by re-estimating the primary hypothesis with the underlying continuous experience variable, which reproduced the effect (OR 1.22 per year, [1.13, 1.33]); continuous or multi-tier classifications might nonetheless refine this measurement in future work.

A third concern is that our outcome measures articulation, not competence. A CV that does not mention leadership is evidence about the document, not about the candidate. All claims in this paper are, therefore, claims about self-presentation, where we discuss the underlying competencies, we flag the inference explicitly as an interpretation that our data support but do not establish.

\subsection{Internal Validity}

A primary internal validity threat concerns extraction noise. The optimized Gemini 3 Pro Preview configuration achieved a precision of 0.69 and a recall of 0.74 on the human-annotated evaluation set, meaning that approximately 30\% of extracted skills lack direct textual grounding and approximately 25\% of genuinely present skills are missed. The reported prevalences should therefore be read as estimates rather than exact measurements. Section~\ref{sec:robustness} quantifies the consequence rather than merely acknowledging it. Under non-differential misclassification, the correction moves every confirmatory contrast away from the null, and eliminating the primary effect would require differential sensitivity between senior and junior CVs so extreme (0.74 vs.\ 0.36, or above 1.0 in the other direction) that it is not attainable. The comparative claims also rely on relative rather than absolute frequencies, so they are preserved under any extraction error that is approximately uniform across groups.

Furthermore, a threat arises from annotator subjectivity in the construction of the ground truth dataset. The 100-CV evaluation sample was annotated by two human researchers following the TaxoSoft taxonomy, with disagreements resolved by the explicit rules described in Section~\ref{sec:validation-pipeline}; we did not compute a formal inter-coder reliability coefficient, which limits how precisely the reliability of the gold standard can be characterized. Moreover, a threat concerns the LLM-as-a-judge step in the role disambiguation procedure. For 13 CVs in which candidates self-identified as both data scientists and ML engineers, Gemini 2.5 Pro acted as a third annotator alongside the two human researchers. While the LLM aligned with the final majority decision in 11 of 13 cases (84.6\%), generative models may inherit stereotypes from their training data about role boundaries that human annotators do not share. We mitigated this risk through the tri-party majority-voting design, in which the LLM never acted as the sole arbiter and no case was decided without agreement from at least one human annotator. Because the 13 affected CVs represent 4.3\% of the corpus and are confined to the ML--data-science boundary, any residual misassignment would attenuate rather than create the role contrasts between those two groups, and cannot affect the seniority results at all.

\subsection{Conclusion Validity}

Four threats specific to the inferential analysis warrant discussion. First, the hypotheses in Section~\ref{sec:hypotheses} were derived from prior literature but were not preregistered, and they were formulated by researchers who had already constructed the corpus. We mitigate this by fixing the comparison and the predicted direction for each hypothesis before testing, by controlling the family-wise error rate across the confirmatory family, and by reporting the exploratory contrasts separately under false-discovery control. Readers should nonetheless treat the confirmatory label as denoting derivation from prior theory rather than the stronger guarantee that preregistration provides.

Second, the study has limited power for small effects. At $\alpha=0.05$ and 80\% power, the design detects differences of roughly 16--19 percentage points depending on the contrast (Section~\ref{sec:stats}). Non-significant results in this study, particularly the ML engineering narrative-only contrast in H\textsubscript{3b} ($p=0.083$) and the homogeneity tests in H\textsubscript{2f}, should therefore not be read as establishing the absence of an effect. The homogeneity conclusion in H\textsubscript{2f} is best stated as the absence of evidence for role-specific seniority slopes given a sample that could only have detected large ones.

Third, several confirmatory results sit close to the corrected significance threshold. Five of the supported hypotheses (H\textsubscript{1b}, H\textsubscript{1c}, H\textsubscript{2b}, H\textsubscript{2c}, H\textsubscript{2e}), together with the refuted H\textsubscript{1e}, have Holm-adjusted $p$-values of 0.029, and H\textsubscript{2c} additionally has a wide odds-ratio interval reflecting the rarity of strategic thinking in junior CVs (6 occurrences). These should be regarded as the weakest members of the confirmatory set, and they are exactly the results we would expect to be most fragile under replication. The exploratory scan reinforces this: when the whole taxonomy is tested against seniority without prior direction, only leadership survives correction.

Fourth, the tests treat CVs as independent observations. This is reasonable for a corpus of documents authored by distinct individuals, but the corpus is drawn from two sources collected five years apart, and the source is not fully orthogonal to role and seniority. We cannot rule out that some portion of the role or seniority contrasts reflects period effects in CV-writing convention rather than the constructs of interest. Disentangling the two would require a corpus stratified by collection period within each role, which we identify as a target for replication.

\subsection{External Validity}

A first external validity threat concerns the boundary between ML engineering and data science roles. These roles overlap substantially in many organizations, with title conventions varying across industries and regions. Treating them as mutually exclusive categories may therefore introduce classification noise in the comparative analysis. Our tri-party disambiguation procedure addresses this for explicitly ambiguous CVs but does not eliminate borderline cases where role identity is implicit rather than self-declared.

Another threat is dataset composition. The 300-CV corpus is constructed from two sources: a publicly available GitHub corpus collected in 2020~\cite{Jiechieu_2020} and a contemporary supplementary set collected through Google queries between November and December 2025. Both sources have their own selection biases. The GitHub corpus reflects whatever distribution of professionals chose to publish CVs on a public platform around 2020, while the contemporary set reflects candidates whose CVs were sufficiently discoverable through public web search. Both subsets are biased toward candidates who have actively chosen to make their CVs publicly accessible, which is itself a self-selection effect that may not represent the broader workforce.

In addition, another threat is the dataset's English-only nature. Non-English CVs were excluded during filtering, and publicly sourced web CVs are likely to overrepresent regions and conventions where publishing CVs online is common practice. Soft-skill articulation is shaped by cultural conventions of professional self-presentation, and our findings may not generalize to labor markets where different rhetorical norms prevail. Cross-cultural extensions of the analysis are a natural direction for future work.

The final external threat is sample size. The 300 CV corpus, while balanced across the three target roles, captures only a small slice of the global technical workforce and cannot resolve small effects. As described in Section~\ref{sec:methodology}, the choice of 100 CVs per role reflects an upper bound on what was practically feasible to curate at the quality level required for this study; Section~\ref{sec:stats} states precisely which effect sizes this permits us to detect.

\subsection{Reliability}

The reliability threat concerns model availability. The extraction pipeline relies on Gemini 3 Pro Preview, a Preview model whose parameters, behavior, or availability may change without notice. Future replications using the same prompt may obtain different results because the underlying model has been updated, deprecated, or replaced. To decouple the statistical findings from this dependency, the replication package includes the per-CV outcome matrix produced by the pipeline alongside the analysis script, so that every test reported in Section~\ref{sec:results} can be reproduced exactly without re-running the extraction.

\section{Conclusion}
\label{sec:conclusion}

This paper investigated how soft skills are expressed in technical CVs through an empirical study of 300 CVs from software engineers, data scientists, and ML engineers, enabled by a generative extraction pipeline that identifies both explicit and implicit representations of behavioral competencies. Rather than reporting frequencies alone, we translated the claims of the demand-side literature into 13 falsifiable hypotheses and tested them inferentially with effect sizes, confidence intervals, and family-wise error control, complemented by false-discovery-controlled exploratory contrasts and by sensitivity analyses that quantify how much extraction error each conclusion could withstand. Eleven hypotheses were supported, one partially, and one refuted.

Four findings stand out. First, role signatures on the candidate side largely mirror the role descriptions in the demand-side literature. Data scientists articulate communication (OR 3.60) and problem solving (OR 2.90) more than the other roles, ML engineers articulate mentoring more (OR 2.11), and data scientists are least likely to produce a CV with no behavioral content at all. Second, the expectation that leadership would be a role-invariant baseline is refuted. Software engineers articulate it at roughly half the rate of the other two roles, and the gap survives adjustment for seniority. Third, soft-skill articulation grows with career progression, with the odds of articulating leadership roughly tripling from junior to senior professionals (adjusted OR 3.06 [1.80, 5.23]); this effect is statistically homogeneous across the three roles rather than concentrated among ML engineers as the raw percentages suggest, and collaboration accumulates alongside leadership rather than being displaced by it. Fourth, candidates convey soft skills predominantly through narrative, with 1,007 implicit skill markers against 347 explicit mentions, and this narrative dominance is concentrated in exactly the competencies that discriminate between candidates. Leadership, coordination, mentoring, and coaching are conveyed through narrative in 88--100\% of cases, while the competencies candidates do list as keywords are the generic ones. Overall, the study suggests that the much-discussed soft skills gap in ML-related professions is, at least in part, an articulation gap. The competencies are present in candidate narratives but remain invisible to keyword-based screening, and invisible in a way that is systematically biased against the most discriminating evidence.

Several directions for future work remain. First, the analysis could be extended beyond CVs to professional profiles on platforms such as LinkedIn, enabling comparisons of self-presentation across document types. Second, cross-cultural and multilingual corpora would test whether the patterns observed here generalize beyond English-language CVs. Third, a preregistered replication on a corpus stratified by collection period would separate role and seniority effects from period effects in CV-writing convention, and would convert the hypotheses tested here into genuinely confirmatory tests. Fourth, on the methodological side, the extraction pipeline could be strengthened by benchmarking emerging model architectures and exploring retrieval-augmented generation (RAG) to improve robustness, and evaluated for whether its recovery of narrative evidence is equally effective across candidate groups.

\section*{CRediT authorship contribution statement}

\textbf{Aidin Azamnouri:} Writing -- original draft, Writing -- review \& editing, Visualization, Validation, Methodology, Investigation, Formal analysis, Data curation, Conceptualization, Supervision. \textbf{Nouran Ayad:} Writing -- review \& editing, Visualization, Validation, Methodology, Investigation, Formal analysis, Data curation. \textbf{Justus Bogner:} Writing -- review \& editing. \textbf{Stefan Wagner:} Writing -- review \& editing, Funding acquisition.

\section*{Statement on open data and ethics}

All artifacts containing the approach implementation, full extraction outputs, prompts, codes, the per-CV outcome matrix, and the complete statistical analysis script are available at \url{https://doi.org/10.5281/zenodo.21873447}.

All CVs analyzed in this study originate from publicly accessible sources. The corpus-derived subset is taken from the multi-labeled resume corpus released by Jiechieu and Tsopze~\cite{Jiechieu_2020}, which was published as a research artifact for academic NLP tasks and is openly available on GitHub. The contemporary subset consists of CVs that the candidates themselves made publicly available on the web, retrieved through targeted web queries restricted to documents the authors explicitly chose to publish in indexed and freely accessible form. Because both sources are already in the public domain, no personal data was recollected, reidentified, or republished as part of this study. The analysis uses these CVs solely for aggregate statistical purposes. All findings reported in this paper are presented at the role or cohort level, and no individual candidate is named, profiled, or identified in the paper, the figures, the tables, or any artifact released alongside it. Any verbatim text used to illustrate examples (e.g., Fig.~\ref{fig2}) shows only short excerpts that do not contain identifying information such as personal names, email addresses, employer names, or phone numbers; these elements were redacted before inclusion in the figure. We do not redistribute the CVs themselves; the replication package consists of the extraction outputs (skill labels) along with code and prompts. A subset of contemporary CVs also includes explicit author statements stating that the document should not be redistributed or republished, which we honor without exception. We further note that anonymizing publicly indexed CVs before redistribution would offer no meaningful privacy protection, because the originals remain freely accessible through the same web channels through which they were collected; any redacted derivative would still be traceable back to its source via search engine retrieval of the unmodified original. We therefore consider analysis-only use, with no redistribution of the underlying CVs, the most defensible ethical position for this work. This approach is consistent with established practice in empirical software engineering research that uses publicly available web data as a source of analysis~\cite{Wohlin_2024}, and we argue that the aggregate, comparative nature of the analysis poses minimal additional privacy risk beyond what already follows from the public availability of the source documents.

It is also worth noting that the LLMs employed in this study are designed to comply with stringent data protection requirements by ensuring that sensitive conversational data is neither retained nor transmitted to any external servers.

\section*{Declaration of competing interest}

The authors declare that they have no known competing financial interests or personal relationships that could have appeared to influence the work reported in this paper.

\section*{Acknowledgements}

The project on which this work is based was sponsored by the Federal Ministry of Education and Research (BMBF) under the funding code 01IS23069.

\bibliographystyle{cas-model2-names}
\bibliography{cas-refs}

\end{document}